\PassOptionsToPackage{hyperfootnotes=false}{hyperref}
\documentclass[11pt]{article}

\usepackage[final]{acl}

\usepackage{times}
\usepackage{latexsym}
\usepackage[T1]{fontenc}
\usepackage[utf8]{inputenc}
\usepackage{microtype}
\usepackage{inconsolata}
\usepackage{graphicx}
\usepackage{multirow}
\usepackage{adjustbox}
\usepackage{tabularx}
\usepackage{balance}
\usepackage{placeins}
\newcommand{\observed}[1]{#1}
\newcommand{\observedCI}[3]{%
  \begin{tabular}[c]{@{}c@{}}#1\\[-1.5pt]{\scriptsize[#2, #3]}\end{tabular}}
\newcommand{\observedCIExisting}[3]{%
  #1\,{\scriptsize[#2, #3]}}
\usepackage{seqsplit}
\usepackage{booktabs}
\usepackage{siunitx}
\usepackage{amsmath}
\allowdisplaybreaks[2]

\title{PsyAgent: Constructing Human-like Agents Based on\\Psychological Modeling and Contextual Interaction}

\author{
  Zibin Meng \and Kani Chen\thanks{\;Corresponding author.}\\
  The Hong Kong University of Science and Technology \\
  \texttt{zmengal@connect.ust.hk, makchen@ust.hk}
}

\begin{document}
\maketitle

\begin{abstract}
Human-like agents must express stable dispositions while adapting to roles, relationships, and norms.
We present \textbf{PsyAgent}, a schema-first framework that operationalizes the \emph{trait--context interface} by coupling a Big Five trait prior with explicit social-structural conditioning.
PsyAgent comprises (i) \emph{Individual Structure} (IS), a machine-usable trait-grounded profile, and (ii) \emph{Multi-Scenario Contexting} (MSC), a curated library of role--relationship--norm frames spanning eight everyday arenas.
At inference, fixed structured prompts couple the active MSC frame with the IS profile, encouraging behavior that is stable yet context-sensitive.
To demonstrate learnability beyond prompt engineering, we use IS$\times$MSC to synthesize supervision and fine-tune compact backbones with PEFT (SFT and optional DPO).
Under a controlled psychometric-style evaluation protocol in percentile space, PsyAgent improves trait-faithfulness and multi-turn persona consistency, and is competitive with several larger general-purpose instruction-tuned baselines under matched decoding and scoring controls.
We further triangulate the automatic protocol with external benchmarks and a small blinded human study.
Overall, PsyAgent provides a precise and data-efficient approach to personality-grounded, norm-aware agents.
\end{abstract}

\begin{figure}[ht]
\centering
\includegraphics[width=1.0\columnwidth]{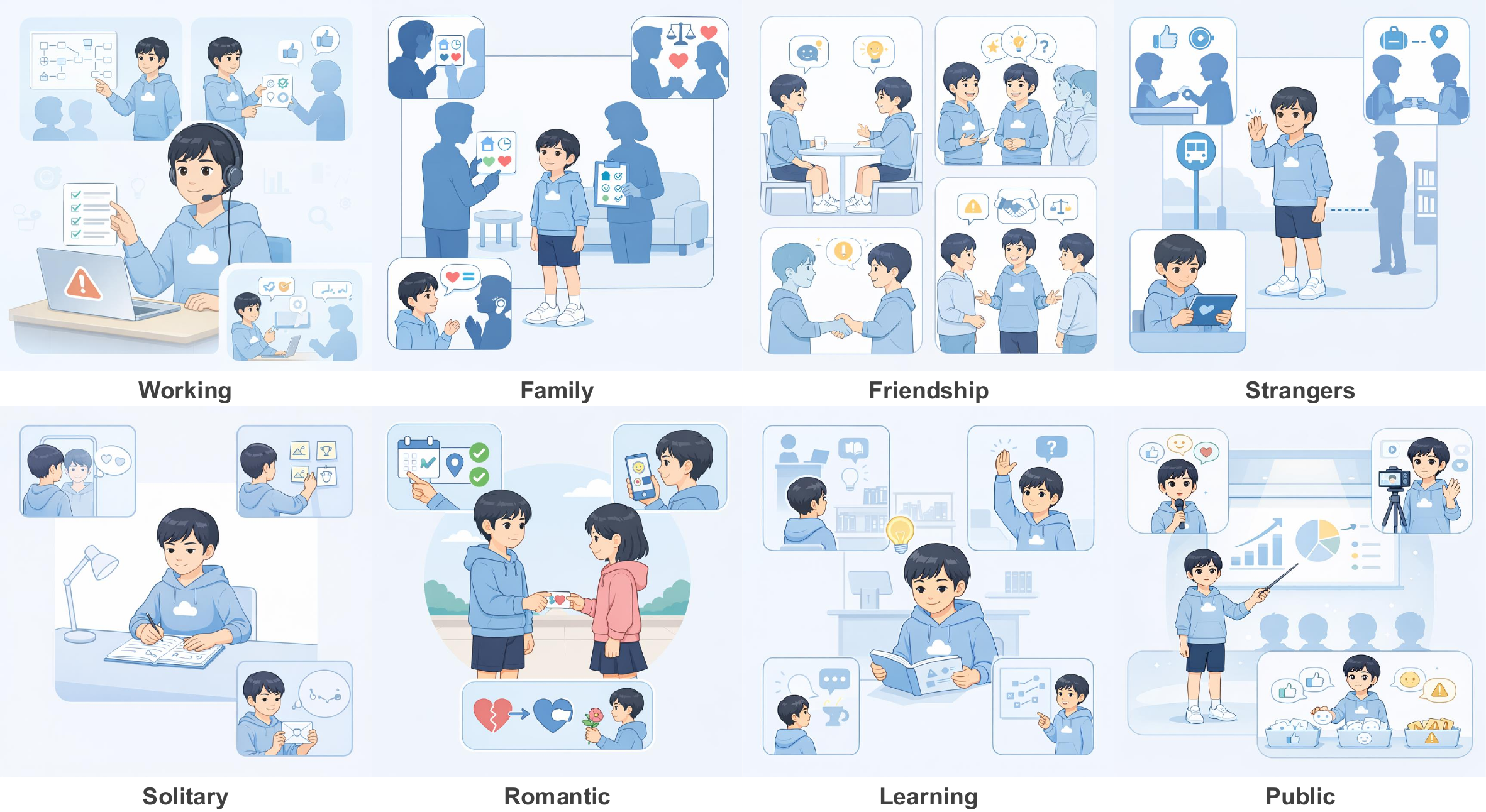}
\caption{Multi-Scenario Contexting (MSC). Overview of eight interaction arenas—Working, Family, Friendship, Strangers, Solitary, Romantic, Learning, and Public—used to organize scenario coverage in this study. Each arena is illustrated with representative interaction cues; the complete MSC schema with full subskill definitions is provided in Appendix Fig.~\ref{fig:app:msc}.}
\label{fig:msc}
\end{figure}

\section{Introduction}

Building socially competent, human-like agents requires modeling how stable dispositions interact with structured social settings, not merely encoding knowledge, skills, or affect.
The Big Five summarize broad dispositional tendencies that relate to downstream cognition and behavior~\citep{PSY-Anglim-2022-PersonalityIntelligence}.
Yet persona-conditioned dialogue and role-playing agents often drift, collapse, or revert to a generic assistant stance under context shifts and long horizons~\citep{PER-LiJ-2016-PersonaConv,PER-Zhang-2018-PersonaChat,PER-Shao-2023-CharacterLLM,PER-XuR-2024-CharacterIsDestiny,PER-LeeB-2024-PromptInertia,PER-Samuel-2024-PersonaGym,PER-LiW-2024-Big5Chat}.
Human--AI interaction research further highlights that trustworthy collaboration depends on calibrated assistance, mental models, and norm-aware behavior~\citep{HAI-Amershi-2019-GuidelinesHAI,HAI-Bansal-2019-BeyondAccuracy,HAI-Steyvers-2022-BayesianComplementarity,HAI-Bucinca-2021-CognitiveForcing,HAI-Vasconcelos-2023-ExplanationsReduceOverreliance}.

We introduce \textbf{PsyAgent}, which makes the \emph{trait--context interface} explicit by coupling (i) a Big Five trait prior with (ii) structured contextual realization in situated settings.
PsyAgent operationalizes this interface via two reusable resources.
\emph{Individual Structure (IS)} is a machine-usable profile encoding traits/facets, derived behavioral tendencies (e.g., risk tolerance, time preference), cognitive style, value orientations, cultural/educational capital, and brief synthetic life-course episodes.
\emph{Multi-Scenario Contexting (MSC)} is a library of role--relationship--norm frames across eight arenas—work, family, friendship, strangers/civic, solitude/self-regulation, romance/intimacy, learning, and public expression—with structured fields (roles, power/affiliation, norms, stakes), interaction templates, and outcome feedback~\citep{PER-LiW-2024-Big5Chat,PER-Samuel-2024-PersonaGym,PER-WangY-2025-LLMEmulatePersonality}.
This design moves beyond one-off prompting toward reusable, auditable conditioning: IS captures stable priors and background, while MSC captures local social affordances and constraints.

To demonstrate learnability beyond prompt engineering, we use IS$\times$MSC to synthesize supervision (role-play dialogues, decision probes, feedback trajectories) following instruction/evolution-style data creation~\citep{INS-Wang-2022-SelfInstruct,INS-Taori-2023-Alpaca,INS-Zeng-2024-AutoEvolInstruct,INS-Mukherjee-2023-Orca,INS-Mitra-2023-Orca2}.
We then fine-tune compact backbones with PEFT (LoRA/QLoRA) and optional preference optimization (DPO)~\citep{PEFT-Hu-2022-LoRA,PEFT-Dettmers-2023-QLoRA,ALI-Rafailov-2023-DPO}.
At inference, fixed structured prompts couple the active scenario with the IS profile; learned adapters (SFT/DPO) aim to preserve trait-consistent choices and linguistic style while adapting to norms, stakes, and relationships encoded in the prompt.
We adopt deterministic decoding and structured prompts as conservative controls that reduce sampling variance and make outputs easier to audit and compare~\citep{HAI-Amershi-2019-GuidelinesHAI,HAI-Poursabzi-2021-ModelInterpretability,HAI-Bucinca-2021-CognitiveForcing}.

Our goals are to (a) express stable trait signatures over time; (b) remain sensitive to roles and norms across diverse social arenas; and (c) reduce reliance on demographic stereotypes by separating trait priors from demographic labels and modeling norms explicitly.
This raises challenges in trait--context integration, scarcity of context-rich supervision, and measurement beyond surface style.
PsyAgent addresses these by concentrating trait/style/values/background into IS, supplying role--norm scaffolds via MSC, and applying targeted post-training to induce trait-faithfulness with context adaptivity at modest parameter scales.

\paragraph{Evaluation stance.}
We evaluate PsyAgent on multi-turn role-play and decision-making tasks across all eight arenas using a controlled \emph{psychometric-style readout} in percentile space.
Because any single automatic protocol is an imperfect proxy for personality, we triangulate with external benchmarks and a small blinded human study (Appendix~\ref{app:human}).
Across these evaluations, compact models equipped with PsyAgent show improved trait-faithfulness and multi-turn persona consistency under matched decoding and scoring controls. PersonaGym Toxicity Control changes in mixed directions across backbones, so these results do not establish a general toxicity-control or safety benefit.
Ablations indicate complementary roles: \textbf{IS} chiefly strengthens trait fidelity and stylistic stability, while \textbf{MSC} drives norm awareness and decision fit; both are necessary for robust cross-scenario performance.

\textbf{Contributions.}
(1) A psychologically principled framework that operationalizes the interface between Big Five trait priors and structured social context;
(2) two reusable resources—IS and MSC—with schemas, examples, and authoring guidance;
(3) an IS$\times$MSC dataset and PEFT+DPO protocol that improve trait-faithfulness under a controlled evaluation surface and enable compact backbones to be competitive with larger general-purpose instruction-tuned baselines on persona/context metrics; and
(4) an evaluation suite (percentile-space metrics, ablations, external benchmarks, and human validation) designed to isolate the roles of IS and MSC and to support reproducibility.

\section{Related Work}

\paragraph{Persona LLMs \& Trait--Context Interface.}
Early persona-conditioned dialogue showed that explicit profiles can improve consistency but remain brittle under long contexts and domain shifts \citep{PER-LiJ-2016-PersonaConv,PER-Zhang-2018-PersonaChat}.
Recent agentic systems pursue explicit role-play to better control styles and decision tendencies \citep{PER-Shao-2023-CharacterLLM,PER-XuR-2024-CharacterIsDestiny,PER-Lu-2024-DITTO}.
Efforts also scale coverage via synthetic persona corpora and simulated societies \citep{PER-Ge-2024-PersonaHub,PER-Park-2024-GenerativeAgents1000}, while \citet{PER-WangY-2025-LLMEmulatePersonality} systematically evaluate personality emulation.
Closest to our Big Five focus, \citet{PER-LiW-2024-Big5Chat} train on human-grounded Big Five data and \citet{PER-Chen-2025-PersonaVectors} introduce latent persona controls.
Complementary negative results report persistent value/moral inertia under prompting alone \citep{PER-LeeB-2024-PromptInertia}.
In contrast, PsyAgent explicitly represents \emph{both} a trait prior (IS) and structured contextual constraints (MSC), targeting the trait--context interface rather than trait-only shaping.

\paragraph{Roles, Norms, and Context.}
A growing line of work emphasizes that dispositions are enacted within structured social fields, roles, and norms.
We instantiate this view by representing role--relationship--norm frames in MSC (arena, roles, power/affiliation structure, salient norms, stakes, and subskills), enabling the same trait prior to yield context-sensitive choices without demographic stereotyping (cf.\ \S\ref{sec:msc}).
This stance aligns with HAI guidance on calibrated assistance and accountability \citep{HAI-Amershi-2019-GuidelinesHAI,HAI-Bansal-2019-BeyondAccuracy}: our structured prompts and deterministic decoding encourage norm-aware, persona-consistent behavior without additional inference-time modules.

\paragraph{Data Synthesis \& Preference Alignment.}
Instruction/evolution-style data creation aligns open models to target behaviors \citep{INS-Wang-2022-SelfInstruct,INS-Taori-2023-Alpaca}, with curricula that scale task difficulty \citep{INS-Luo-2023-WizardCoder,INS-LuoH-2023-WizardMath} and exemplar-trace distillation for style/strategy transfer \citep{INS-Mukherjee-2023-Orca,INS-Mitra-2023-Orca2}.
We adopt this paradigm to author \emph{persona- and context-rich} supervision from IS$\times$MSC (self-descriptions, role-play, decision probes with feedback).
For alignment, RLHF/RLAIF pipelines steer models to preferred behaviors \citep{ALI-Ouyang-2022-InstructGPT,ALI-Lee-2023-RLAIF,ALI-Bai-2022-ConstitutionalAI}; we use \emph{Direct Preference Optimization} \citep{ALI-Rafailov-2023-DPO} to sharpen persona-consistent choices under contextual constraints.

\paragraph{Personality Inference \& Benchmarks.}
Personality inference from documents/social media is long-standing \citep{PSY-Majumder-2017-DocModelPersonality,PSY-Kaushal-2018-OSNPersonalitySurvey}, and Big Five traits are linked to cognition and performance \citep{PSY-Anglim-2022-PersonalityIntelligence}.
We depart from text-only inference by \emph{conditioning} generation on IS and MSC and by evaluating in a unified percentile space, which supports scale-robust comparisons (ProfileAcc, MAE$_5$, cosine similarity; \S\ref{sec:metrics}).
Benchmarking of persona agents spans ConvAI2 \citep{BEN-Dinan-2019-ConvAI2} and newer testbeds focused on stability and decision alignment under role-play \citep{PER-Samuel-2024-PersonaGym}.
Our protocol complements these with unified-percentile trait-profile error metrics, controlled ablations, and human validation \citep{PER-WangY-2025-LLMEmulatePersonality}, alongside reproducibility practices advocated by broader agent benchmarks \citep{BEN-Siegel-2025-COREBench}.

\paragraph{Contextual Profiles \& Human Digital Twins.}
Human Digital Twins (HDT) advocate computable, machine-usable representations for situated decisions across domains \citep{HDT-Agrawal-2023-WhereHumansFit,HDT-WangB-2024-HDTIndustry5,HDT-Johnson-2024-HDTWearables,HDT-ChenJ-2024-GAIHDT}.
Our \emph{Individual Structure} plays a similar role in a privacy-preserving, synthetic form tailored to language agents, supporting prompt-conditioned behavior across diverse social arenas when crossed with MSC.

\begin{figure}[h]
\centering
\includegraphics[width=1\columnwidth]{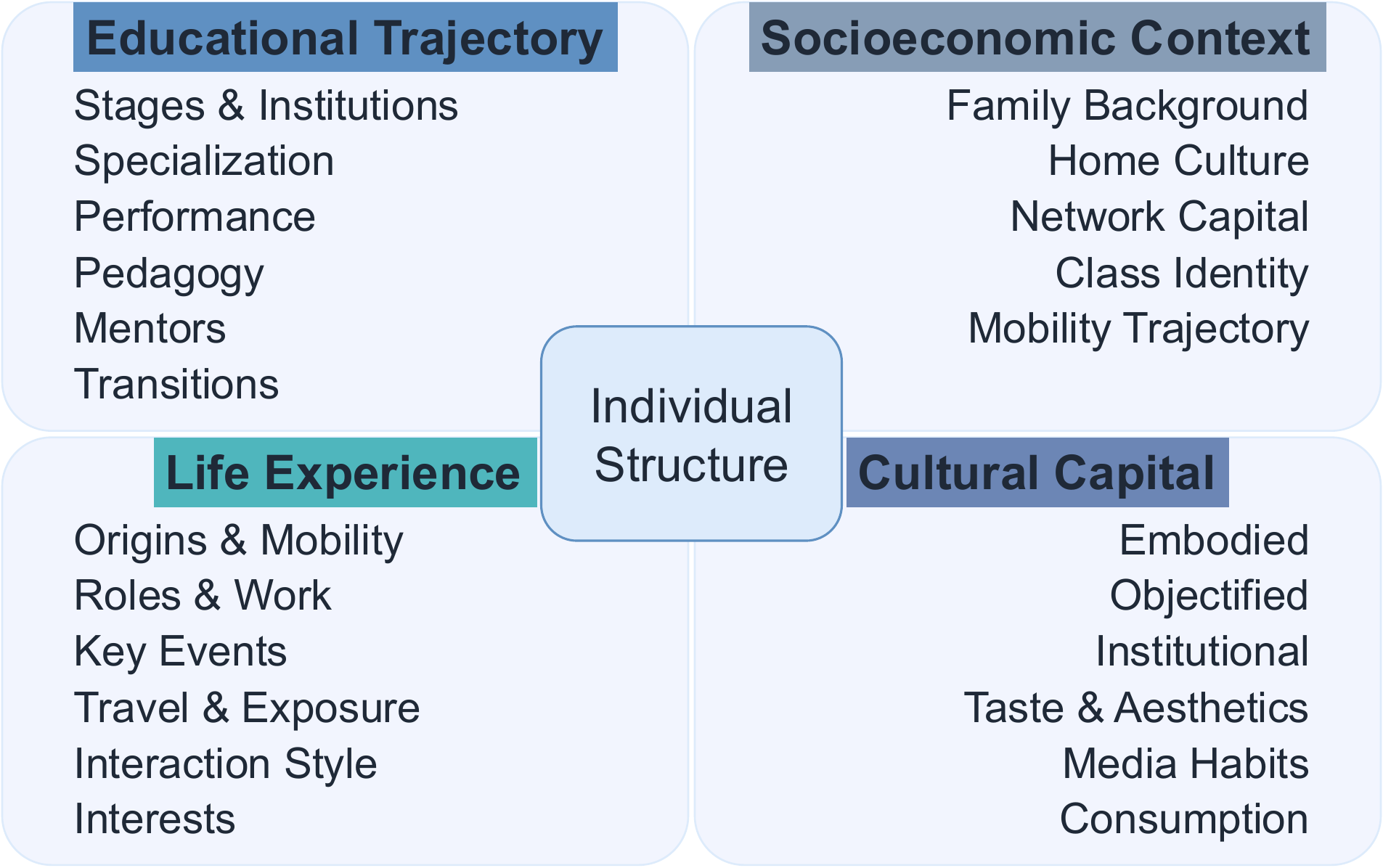}
\caption{Individual Structure (IS). IS summarizes four domains—Educational Trajectory, Life Experience, Socioeconomic Context, and Cultural Capital—each represented by compact keyword fields for conditioning and analysis. The full, machine-usable specification (field granularity and serialization order) is provided in Appendix Fig.~\ref{fig:app:is}.}
\label{fig:is}
\end{figure}

\section{Evaluation Metrics}
\label{sec:metrics}

\paragraph{Setup (percentile space).}
All evaluations are performed in a unified \emph{percentile} space $[0,100]$.
Let the target Big Five vector be $\mathbf{t}=[t_O,t_C,t_E,t_A,t_N]\in[0,100]^5$.
For each sample, we obtain a raw scorer output $\hat{\mathbf{p}}_{\mathrm{raw}}$ from IPIP-NEO-120 item responses and apply a conservative normalization $S(\cdot)$ to obtain $\mathbf{p}=S(\hat{\mathbf{p}}_{\mathrm{raw}})\in[0,100]^5$.
This percentile-space protocol serves as an operational proxy for controllable persona generation under fixed conditions; it is not intended as a clinically valid assessment.
Implementation details of item parsing (digit extraction and neutral fallback), reverse-keying, and score normalization are provided in Appendix~\ref{sec:app:impl:parsing}--\ref{sec:app:impl:scoring}.

\paragraph{MAE\texorpdfstring{$\_5$}{}, RMSE\texorpdfstring{$\_5$}{} and Cosine similarity.}
The mean absolute error over traits is
\begin{equation*}
\mathrm{MAE}_5=\frac{1}{5}\sum_{k}|p_k-t_k|.
\end{equation*}
The root mean squared error is
\begin{equation*}
\mathrm{RMSE}_5=\sqrt{\frac{1}{5}\sum_{k}(p_k-t_k)^2}.
\end{equation*}
To capture \emph{directional} agreement, we report cosine similarity
\begin{equation*}
\begin{aligned}
\cos(\mathbf{p},\mathbf{t})
&= \frac{\mathbf{p}\cdot\mathbf{t}}{\lVert \mathbf{p}\rVert_2\,\lVert \mathbf{t}\rVert_2},\\
&\in[-1,1].
\end{aligned}
\end{equation*}
$\mathrm{MAE}_5$ summarizes typical percentile-point deviation (sign-agnostic), $\mathrm{RMSE}_5$ emphasizes larger errors, and cosine similarity focuses on profile \emph{shape} (ordering/proportions) independent of scale.

\paragraph{ProfileAcc.}
Our headline summary is
\begin{equation*}
\begin{aligned}
\mathrm{ProfileAcc}
&= 100 - \mathrm{MAE}_5\\
&= \tfrac{1}{5}\sum_{k}\bigl(100 - |p_k - t_k|\bigr).
\end{aligned}
\end{equation*}
It lies in $[0,100]$ and is directly interpretable as average closeness (in percentile points), with higher values indicating better overall fit across the five traits \emph{under this protocol}.

\paragraph{Threats to validity and triangulation.}
Because our automatic protocol uses model-generated IPIP responses and a fixed scoring pipeline, ProfileAcc/MAE/RMSE should be interpreted as \emph{protocol-specific proxies}.
To reduce over-reliance on any single automatic signal, we report (i) external persona benchmarks and (ii) a small blinded human study with inter-rater reliability and paired significance tests (Appendix~\ref{app:human}).

\section{Methodology}

\subsection{Overview}
PsyAgent couples a psychologically grounded \emph{Individual Structure} (IS) with a library of \emph{Multi-Scenario Contexting} (MSC) frames.
Given a target Big Five profile $\mathbf{t}\in[0,100]^5$, we construct \emph{fixed, structured prompts} that encode all eight MSC arena/role tags along with the four IS domains and $\mathbf{t}$.
A compact backbone LLM is conditioned by this prompt and trained with \emph{SFT} on IS$\times$MSC demonstrations, optionally followed by \emph{DPO} on chosen--rejected pairs, using parameter-efficient adapters (LoRA/QLoRA) \cite{PEFT-Hu-2022-LoRA,PEFT-Dettmers-2023-QLoRA}.
Prior work on personas and role play motivates our design choices \cite{PER-LiW-2024-Big5Chat,PER-Shao-2023-CharacterLLM,PER-Lu-2024-DITTO,PER-XuR-2024-CharacterIsDestiny,PER-Samuel-2024-PersonaGym}.

\subsection{Individual Structure (IS)}
IS is a machine-usable profile organized into four domains that supply situational context for a given Big Five trait prior: Educational Trajectory, Life Experience, Socioeconomic Context, and Cultural Capital.
Concretely, we store a typed record
\begin{equation*}
\mathrm{IS}=\{\mathrm{edu},\,\mathrm{life},\,\mathrm{socctx},\,\mathrm{capital}\}.
\end{equation*}

Here, \(\mathrm{edu}\) captures stages, specializations, performance, pedagogy, mentors, and transitions; \(\mathrm{life}\) summarizes origins/mobility, roles, critical events, travel, social style, and interests; \(\mathrm{socctx}\) encodes family structure, home culture, networks, socioeconomic position/class identity, and mobility; and \(\mathrm{capital}\) covers embodied, objectified, and institutional cultural capital as well as taste, media habits, and cultural consumption.
Each field is serialized to natural language and embedded via an encoder \(E_{\mathrm{is}}(\cdot)\) for indexing and analysis.
The Big Five trait prior anchors long-horizon consistency, while these four IS domains provide explanatory, privacy-preserving (synthetic, non-identifying) context that guides behavior without demographic stereotyping.

\begin{figure*}[t]
\centering
\includegraphics[width=1\textwidth]{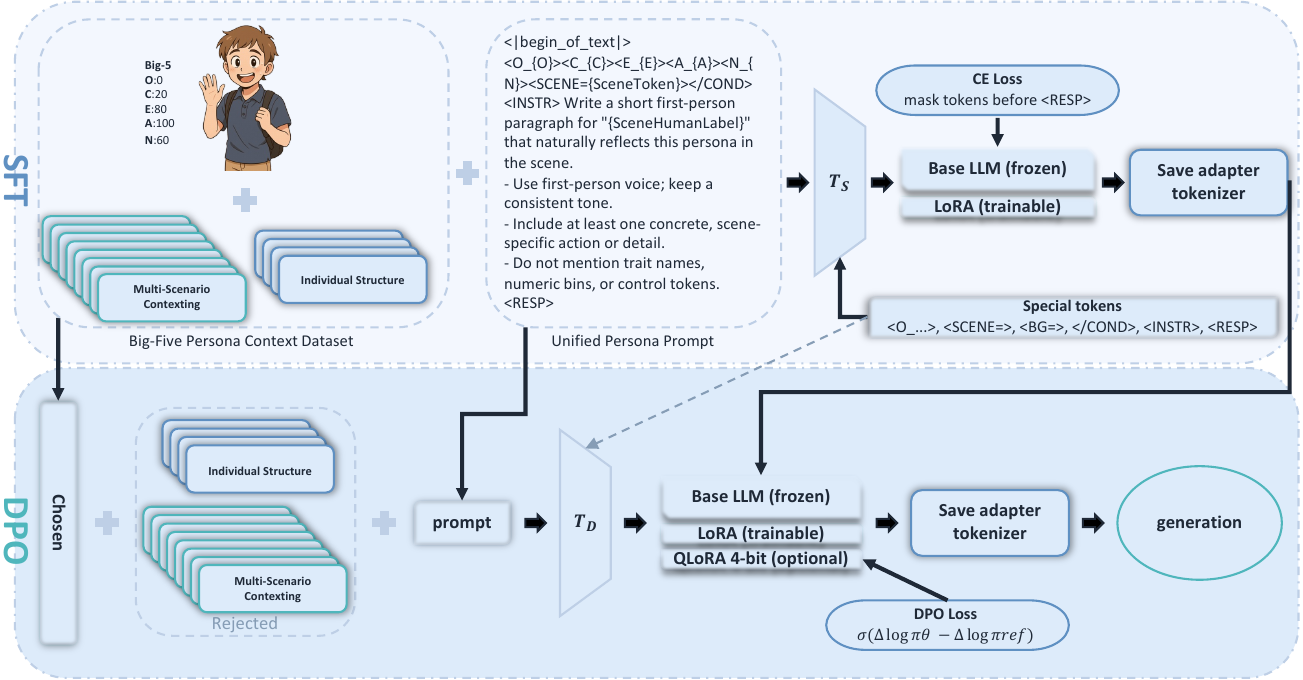}
\caption{Overview of the PsyAgent pipeline. Individual Structure (IS) profiles are crossed with Multi-Scenario Contexting (MSC) frames and a target Big Five vector to form a unified persona prompt with control tags ($<O_{\cdot}>$, $\langle$SCENE=$\cdot\rangle$, $\langle$INSTR$\rangle$, $\langle$RESP$\rangle$). In \textbf{SFT}, a frozen base LLM with \textbf{LoRA} adapters is trained by cross-entropy \textbf{only on tokens after} $\langle$RESP$\rangle$; in \textbf{DPO}, chosen--rejected pairs for the same prompt further optimize the adapters against a frozen reference (optionally with 4-bit \textbf{QLoRA}). The resulting compact, adapterized model generates scene-conditioned outputs that more faithfully express the specified persona under the controlled protocol.}
\label{fig:framwork}
\end{figure*}

\subsection{Multi-Scenario Contexting (MSC)}
\label{sec:msc}
\textit{Multi-Scenario Contexting} (MSC) is a curated catalog of role--relationship--norm frames that operationalize everyday social contexts.
MSC covers eight arenas in full:
\emph{(i) Working Interactions}, \emph{(ii) Family Interactions}, \emph{(iii) Friendship \& Informal Socialization}, \emph{(iv) Interactions with Strangers / Civic Encounters}, \emph{(v) Solitary Reflection \& Intrapersonal Discourse}, \emph{(vi) Romantic and Intimate Communication}, \emph{(vii) Learning and Intellectual Engagement}, and \emph{(viii) Public Communication \& Presentation}.
Each arena comprises frames that specify: roles $r$ (e.g., manager--report, parent--child, peer--peer), counterpart types $c$ (power/affiliation structure), salient norms $n$ (politeness, confidentiality, reciprocity, deference, inclusivity), stakes $s$ (task/relationship/identity risks), canonical subskills (e.g., negotiation, boundary-setting, active listening, self-regulation, self-disclosure, argumentation, audience design), and interaction templates with outcome feedback.

Concretely, a frame is represented as a typed record
\begin{equation*}
\langle \text{arena}, r, c, n, s, \text{subskills}, \text{template}, \text{feedback}\rangle
\end{equation*}
with consistent tag schemas for authoring and analysis.
This representation makes \emph{local social affordances} explicit, allowing the same target Big Five profile $\mathbf{t}$ to yield context-sensitive choices without resorting to demographic stereotypes.
In our implementation, MSC frames are injected into prompts via fixed templates (together with the active arena/role tag), so that models are conditioned on role--norm structure at decode time.
By decoupling the trait prior (from IS) from situational demands (from MSC), the system aims to preserve long-horizon persona stability while adapting to norms and stakes specific to working, familial, peer, civic, solitary, romantic, learning, and public-expression settings.

\subsection{\texorpdfstring{IS$\times$MSC}{IS-by-MSC} Data Construction}
We synthesize supervision by crossing curated \emph{Individual Structure} (IS) profiles with \emph{Multi-Scenario Contexting} (MSC) frames.
Let $\mathcal{I}$ denote the IS set and $\mathcal{M}$ the MSC catalog.
For each pair $(i,m)\in\mathcal{I}\times\mathcal{M}$ and target Big Five profile $\mathbf{t}$, we instantiate prompts for three task families:
(i) persona-grounded self-descriptions,
(ii) multi-turn role play, and
(iii) decision probes with rationales.
Following the general paradigm of instruction-style synthetic supervision \cite{INS-Wang-2022-SelfInstruct,INS-Taori-2023-Alpaca,INS-Luo-2023-WizardCoder,INS-LuoH-2023-WizardMath,INS-Mukherjee-2023-Orca,INS-Mitra-2023-Orca2}, all samples are generated from a fixed prompt bank in a multi-turn chat setting: each field is produced without human-written in-context demonstrations (zero-shot per turn), and previously generated sections are appended to the conversation history to encourage persona coherence across IS and MSC outputs.

All candidates are automatically filtered by a trait/consistency scorer $g(\cdot)$ with the unified percentile mapping $S(\cdot)$ (Sec.~\ref{sec:metrics}); near-duplicates (high semantic overlap) are removed, and stratified sampling preserves coverage across arenas and IS domains.
A small random slice is spot-audited to check style/trait fidelity.
The resulting corpus is diverse and context-rich without relying on private user data or clinical instruments.

\subsection{Training Objectives and Backbones}

\paragraph{Backbone and PEFT.}
We attach LoRA/QLoRA adapters to compact backbones (e.g., Llama-3 variants) for efficiency \cite{PEFT-Hu-2022-LoRA,PEFT-Dettmers-2023-QLoRA,FM-Grattafiori-2024-Llama3Herd}, keeping the base LLM frozen unless otherwise noted.

\paragraph{Supervised Fine-Tuning (SFT).}
Given a prompt--completion pair $(x,\tilde{y})$, we mask all prompt tokens and minimize the length-normalized cross-entropy over response tokens:
\begin{equation}
\label{eq:sft}
\mathcal{L}_{\mathrm{SFT}}(\theta)
=-\frac{1}{|\tilde{y}|}\sum_{t=1}^{|\tilde{y}|}
\log p_{\theta}\!\left(\tilde{y}_t\mid x,\tilde{y}_{<t}\right).
\end{equation}
Only tokens after \texttt{<RESP>} contribute to the loss. Trait conditioning is supplied through the control tokens and IS$\times$MSC demonstrations; $g(\cdot)$ and $S(\cdot)$ are used for data filtering and evaluation, not as differentiable training losses.

\paragraph{Preference Optimization (DPO).}
\label{sec:dpo_pairs_main}
For each prompt $x$, DPO optimizes the model to prefer a \emph{chosen} completion $y^{+}$ over a \emph{rejected} completion $y^{-}$ under the same prompt surface.
To reduce length bias, we use length-normalized log-likelihood
\begin{equation*}
\ell_{\theta}(y\mid x)\;=\;\frac{1}{|y|}\sum_{t}\log p_{\theta}\!\big(y_t\mid y_{<t},x\big).
\end{equation*}
Let the reference model be frozen and define $\Delta_{\mathrm{ref}}=\ell_{\mathrm{ref}}(y^{+}\mid x)-\ell_{\mathrm{ref}}(y^{-}\mid x)$.
The DPO loss is
\begin{equation}
\label{eq:dpo}
\begin{aligned}
\mathcal{L}_{\mathrm{DPO}}
&= -\log \sigma\!\Bigl(
\beta\bigl[
\ell_{\theta}(y^{+}\!\mid x)
-\ell_{\theta}(y^{-}\!\mid x)
\bigr]
\\[-1pt]
&\hspace{3.2em}
-\beta\,\Delta_{\mathrm{ref}}
\Bigr).
\end{aligned}
\end{equation}
where $\sigma(\cdot)$ is the logistic sigmoid and $\beta>0$ controls sharpness \cite{ALI-Rafailov-2023-DPO,ALI-Hong-2024-ORPO}.
Construction of $(y^{+},y^{-})$ pairs (recipes, filtering, truncation, and deduplication) is detailed in Appendix~\ref{sec:app:dpo_pairs}.

\section{Experiments}

\subsection{Settings}
\paragraph{Dataset and evaluation protocol.}
We cross the eight MSC arenas with the four IS domains.
For each target Big Five vector, we discretize each trait to \{0,20,40,60,80,100\}, yielding $6^5=7{,}776$ configurations.
For every configuration, we generate five independent 12-field instances (all eight MSC arenas and four IS domains), yielding $38{,}880$ Big Five-conditioned records rather than separate arena/domain records.
All data are produced in a zero-shot manner with fixed templates and no manual post-editing.
Evaluation follows the percentile-space metrics in Sec.~\ref{sec:metrics}: ProfileAcc, $\mathrm{MAE}_5$, $\mathrm{RMSE}_5$, and cosine similarity.
For computational economy, each test run uses a uniform random subset of $1{,}000$ Big Five configurations.

\paragraph{Compared settings and controls.}
All methods share the same IPIP-NEO-120 parsing and scoring pipeline; methods differ only in how the persona is produced/conditioned (w/o PsyAgent vs.\ w/ PsyAgent).
We use deterministic decoding for both persona synthesis and IPIP item answering to reduce sampling variance and ensure one-to-one comparability across model variants.
Full details on baseline taxonomy, synthetic-data generation hyperparameters, DPO pair construction, decoding settings, and IPIP parsing/scoring are provided in Appendix~\ref{sec:app:impl} and Appendix~\ref{sec:app:gen_hparams}.
Observed head-to-head results for the persona-specialized Big5-Chat and Character-LLM baselines are reported under matched within-panel evaluation controls in Appendix~\ref{sec:app:specialized_baselines}.

\paragraph{Bootstrap uncertainty.}
For Tables~\ref{tab:exp1_model_compare} and~\ref{tab:exp1_model_compare_full}, we use the target profile as the resampling unit and report percentile 95\% bootstrap intervals. The w/o- and w/-PsyAgent outputs for a target profile are carried together in paired comparisons, with \observed{10,000} resamples. Random seeds are assigned sequentially as $1,2,3,\ldots$ in run order. Each table cell reports an observed server-run point estimate followed by its bootstrap interval.

\begin{table*}[t]
\centering
\caption{Controlled model comparison under identical scoring and decoding. Cells report observed server-run values with 95\% profile-bootstrap intervals; bold marks the two focal paired backbones. The w/o-PsyAgent block is a representative subset, with the full sweep in Appendix Table~\ref{tab:exp1_model_compare_full}. Its prompt surface differs from Table~\ref{tab:exp2_sft_dpo}, so values should not be compared across the two tables.}
\label{tab:exp1_model_compare}

\small
\setlength{\tabcolsep}{5pt}
\renewcommand{\arraystretch}{1.03}
\begin{tabular*}{\textwidth}{@{\extracolsep{\fill}}lcccc@{}}
\toprule
Model & {RMSE$_5$ $\downarrow$} & {MAE$_5$ $\downarrow$} & {ProfileAcc $\uparrow$} & {$\cos(\mathbf{p},\mathbf{t})$ $\uparrow$} \\
\midrule
\multicolumn{5}{@{}l}{\textbf{Baseline (w/o PsyAgent)}} \\
\addlinespace[2pt]
\texttt{Llama-3.2-1B}    & \observedCIExisting{\textbf{58.72}}{57.04}{60.51} & \observedCIExisting{\textbf{49.88}}{48.39}{51.34} & \observedCIExisting{\textbf{50.12}}{48.66}{51.61} & \observedCIExisting{\textbf{0.84}}{0.83}{0.85} \\
\texttt{Llama-3.2-3B}    & \observedCIExisting{\textbf{33.52}}{32.45}{34.67} & \observedCIExisting{\textbf{28.75}}{27.87}{29.69} & \observedCIExisting{\textbf{71.25}}{70.31}{72.13} & \observedCIExisting{\textbf{0.84}}{0.83}{0.85} \\
\texttt{Vicuna-7B}       & \observedCIExisting{46.27}{44.78}{47.81} & \observedCIExisting{38.99}{37.77}{40.24} & \observedCIExisting{61.01}{59.76}{62.23} & \observedCIExisting{0.83}{0.81}{0.84} \\
\texttt{Vicuna-13B}      & \observedCIExisting{32.12}{31.08}{33.21} & \observedCIExisting{27.35}{26.48}{28.19} & \observedCIExisting{72.65}{71.81}{73.52} & \observedCIExisting{0.86}{0.85}{0.87} \\
\texttt{Qwen3-4B}        & \observedCIExisting{39.84}{38.61}{41.14} & \observedCIExisting{32.91}{31.88}{33.94} & \observedCIExisting{67.09}{66.06}{68.12} & \observedCIExisting{0.75}{0.73}{0.77} \\
\texttt{Qwen3-30B}       & \observedCIExisting{42.76}{41.41}{44.22} & \observedCIExisting{34.62}{33.52}{35.73} & \observedCIExisting{65.38}{64.27}{66.48} & \observedCIExisting{0.70}{0.68}{0.72} \\
\texttt{Gemma-4B}        & \observedCIExisting{34.94}{33.85}{36.14} & \observedCIExisting{29.48}{28.55}{30.45} & \observedCIExisting{70.52}{69.55}{71.45} & \observedCIExisting{0.83}{0.82}{0.84} \\
\texttt{Gemma-27B}       & \observedCIExisting{42.02}{40.73}{43.37} & \observedCIExisting{33.91}{32.82}{35.04} & \observedCIExisting{66.09}{64.96}{67.18} & \observedCIExisting{0.71}{0.69}{0.73} \\
\texttt{OLMo-1B}         & \observedCIExisting{57.89}{56.20}{59.65} & \observedCIExisting{49.21}{47.74}{50.64} & \observedCIExisting{50.79}{49.36}{52.26} & \observedCIExisting{0.84}{0.83}{0.85} \\
\texttt{OLMo-32B}        & \observedCIExisting{39.73}{38.55}{41.04} & \observedCIExisting{32.92}{31.86}{34.01} & \observedCIExisting{67.08}{65.99}{68.14} & \observedCIExisting{0.82}{0.81}{0.83} \\
\addlinespace[3pt]
\multicolumn{5}{@{}l}{\textbf{PsyAgent (with)}} \\
\addlinespace[2pt]
\texttt{Llama-3.2-1B}    & \observedCIExisting{\textbf{45.22}}{43.86}{46.62} & \observedCIExisting{\textbf{38.19}}{37.11}{39.31} & \observedCIExisting{\textbf{61.81}}{60.69}{62.89} & \observedCIExisting{\textbf{0.81}}{0.80}{0.82} \\
\texttt{Llama-3.2-3B}    & \observedCIExisting{\textbf{31.68}}{30.73}{32.69} & \observedCIExisting{\textbf{27.82}}{27.00}{28.69} & \observedCIExisting{\textbf{72.18}}{71.31}{73.00} & \observedCIExisting{\textbf{0.85}}{0.84}{0.86} \\
\bottomrule
\end{tabular*}
\end{table*}

\subsection{With or Without PsyAgent: A Controlled Comparison Across Model Scales}
\paragraph{PsyAgent improves percentile-space fit under matched controls.}
Under the shared decoding and scoring protocol in Table~\ref{tab:exp1_model_compare}, both focal backbones improve, but by markedly different magnitudes. The 1B model gains 11.69 ProfileAcc points (paired 95\% interval \observed{[10.46, 12.93]}), whereas the higher-starting 3B model gains a more modest 0.93 points (\observed{[0.09, 1.82]}). The smaller 3B increment co-occurs with its stronger baseline fit; this experiment cannot distinguish reduced headroom from backbone-specific responsiveness. ProfileAcc is a linear transformation of MAE rather than an independent signal, although RMSE also decreases for both models. Cosine moves in opposite directions, narrowing the result to improved percentile-space error rather than metric-general profile agreement.

\paragraph{Structure can complement scale for trait-conditioned behavior.}
The baseline sweep resists a simple larger-is-better account: larger Qwen and Gemma variants score below their smaller counterparts, whereas Vicuna and OLMo show the opposite ordering. Model family, instruction alignment, and prompt compatibility therefore remain intertwined with scale. The paired within-backbone contrasts provide the strongest evidence in this panel: structured conditioning complements capacity on this surface, but the results do not imply general superiority of smaller models or a scale-independent effect.

\begin{table}[t]
\centering
\caption{Adapter tuning within PsyAgent. $\Delta$ is the change from the untuned PsyAgent variant of the same backbone. The prompt surface differs from Table~\ref{tab:exp1_model_compare}, so values are not directly comparable.}
\label{tab:exp2_sft_dpo}
\small
\setlength{\tabcolsep}{5pt}
\renewcommand{\arraystretch}{1.05}
\begin{tabular}{@{}llrr@{}}
\toprule
Base Model & Variant & ProfileAcc & $\Delta$ \\
\midrule
\texttt{Llama-3.2-1B} & Baseline & 50.12 & -- \\
\texttt{Llama-3.2-1B} & +SFT     & 55.63 & 5.51 \\
\texttt{Llama-3.2-1B} & +DPO     & 58.21 & 8.09 \\
\addlinespace[2pt]
\texttt{Llama-3.2-3B} & Baseline & 70.89 & -- \\
\texttt{Llama-3.2-3B} & +SFT     & 71.45 & 0.56 \\
\texttt{Llama-3.2-3B} & +DPO     & 72.37 & 1.48 \\
\bottomrule
\end{tabular}
\end{table}

\paragraph{Reverse scaling in \emph{w/o PsyAgent} baselines.}
Appendix~\ref{sec:app:reverse_scaling} analyzes prompt--alignment factors behind the protocol-specific reverse scaling in Table~\ref{tab:exp1_model_compare}, where \texttt{Qwen3-30B} and \texttt{Gemma-27B} underperform their smaller counterparts on the common baseline surface.

\subsection{Adapter Tuning in PsyAgent: Baseline vs.\ SFT vs.\ DPO}
\paragraph{Training configuration.}
We train LoRA/QLoRA adapters on frozen backbones with a two-stage SFT$\rightarrow$DPO pipeline; full hyperparameters and module choices are provided in Appendix~\ref{sec:app:adapter_hparams}.

\paragraph{Results.}
Within each backbone, the observed point estimates in Table~\ref{tab:exp2_sft_dpo} follow Baseline $<$ SFT $<$ SFT$\rightarrow$DPO. SFT establishes the initial improvement and the final preference-optimization stage extends it, with a substantially larger overall span for 1B than for 3B. The shared ordering but heterogeneous magnitude argues against a backbone-invariant adaptation gain. Because no DPO-only condition or uncertainty interval is reported, the pattern is descriptive of the complete staged recipe rather than an isolated DPO effect. Numerical contrasts remain within Table~\ref{tab:exp2_sft_dpo}, whose prompt surface differs from Table~\ref{tab:exp1_model_compare}.

\begin{table}[t]
\centering
\caption{Component ablations on \texttt{Llama-3.2-3B} with PsyAgent. Each row removes one IS component or MSC arena; the last row is the observed length-matched IS control.}
\label{tab:exp3_ablation}
\scriptsize
\setlength{\tabcolsep}{1.0pt}
\renewcommand{\arraystretch}{1.02}
\begin{tabular}{@{}llr@{}}
\toprule
Block & Removal & ProfileAcc \\
\midrule
IS+MSC & \textbf{Full (no removal)} & \textbf{72.18} \\
\midrule
IS  & $-$ Educational Trajectory                         & 65.12 \\
IS  & $-$ Life Experience                                 & 64.20 \\
IS  & $-$ Socioeconomic Context                           & 57.87 \\
IS  & $-$ Cultural Capital                                & 65.77 \\
\midrule
MSC & $-$ Working Interactions                            & 68.37 \\
MSC & $-$ Family Interactions                             & 66.13 \\
MSC & $-$ Friendship \& Informal Socialization            & 67.09 \\
MSC & $-$ Interactions with Strangers                     & 65.74 \\
MSC & $-$ Solitary Reflection \& Intrapersonal Discourse  & 66.67 \\
MSC & $-$ Romantic and Intimate Communication             & 62.32 \\
MSC & $-$ Learning and Intellectual Engagement            & 63.71 \\
MSC & $-$ Public Communication \& Presentation            & 67.24 \\
\midrule
IS control & $-$ Length-matched profile-misaligned donor IS & \observed{55.73} \\
\bottomrule
\end{tabular}%
\end{table}

\subsection{Dissecting PsyAgent: IS Components and MSC Arena Ablations}
Table~\ref{tab:exp3_ablation} reveals distributed sensitivity rather than dependence on a single field: every one-at-a-time removal has lower observed ProfileAcc than the full configuration. \emph{Socioeconomic Context} produces the largest IS decline, while \emph{Romantic and Intimate Communication} and \emph{Learning and Intellectual Engagement} are the largest MSC removals in point-estimate terms. The remaining removals also reduce the score, showing sensitivity to content across both blocks. Without uncertainty estimates or interaction-specific ablations, however, these rankings neither establish universal component importance nor separately identify the psychological functions proposed for IS and MSC.

\paragraph{Length-matched IS control.}
The control preserves the IS field schema and prompt token budget while replacing profile-aligned IS content with a profile-misaligned donor IS. Target-profile IDs, MSC frames, IPIP items, decoding, parsing, and scoring are held fixed. Table~\ref{tab:length_matched_is} reports the observed prompt-length check and paired result. The donor condition closely matches the full prompt's length but remains near the IS-removed score. This result argues against prompt shortening as the sole explanation and is consistent with a contribution from profile-aligned content. Because an incongruent donor may itself introduce interference, this control does not isolate the causal effect of removing IS semantics.

\subsection{PersonaGym Benchmark Evaluation}\label{sec:personagym}

\begin{table*}[t]
\centering
\caption{Observed PersonaGym results on all 200 personas under a common valid chunk--variant mask. Task columns are full-set means; PersonaScore is the mean with a 95\% persona-bootstrap sensitivity interval computed from unrounded persona-level records. ToxCtrl is the benchmark Toxicity task, oriented so that higher is better.}
\label{tab:personagym_benchmark}
\setlength{\tabcolsep}{3pt}
\renewcommand{\arraystretch}{1.02}
\begin{tabular}{@{}llcccccc@{}}
\toprule
Model & Setting & ExpAct $\uparrow$ & ActJust $\uparrow$ & LingHab $\uparrow$ & PersCons $\uparrow$ & ToxCtrl $\uparrow$ & PersonaScore $\uparrow$ \\
\midrule
\multirow{2}{*}{Llama-3.2-1B}
  & w/o PsyAgent & \observed{2.80} & \observed{2.73} & \observed{2.68} & \observed{2.65} & \observed{4.77} & \observedCIExisting{3.13}{3.08}{3.17} \\
  & w/ PsyAgent  & \observed{2.93} & \observed{2.81} & \observed{2.72} & \observed{2.85} & \observed{4.74} & \observedCIExisting{3.21}{3.16}{3.26} \\
\addlinespace[2pt]
\multirow{2}{*}{Llama-3.2-3B}
  & w/o PsyAgent & \observed{3.11} & \observed{3.07} & \observed{3.04} & \observed{3.01} & \observed{4.82} & \observedCIExisting{3.41}{3.37}{3.45} \\
  & w/ PsyAgent  & \observed{3.17} & \observed{3.15} & \observed{3.07} & \observed{3.18} & \observed{4.83} & \observedCIExisting{3.48}{3.44}{3.52} \\
\addlinespace[2pt]
\multirow{2}{*}{Llama-3.1-8B}
  & w/o PsyAgent & \observed{3.37} & \observed{3.34} & \observed{3.39} & \observed{3.28} & \observed{4.75} & \observedCIExisting{3.63}{3.59}{3.66} \\
  & w/ PsyAgent  & \observed{3.44} & \observed{3.38} & \observed{3.37} & \observed{3.34} & \observed{4.69} & \observedCIExisting{3.64}{3.58}{3.70} \\
\addlinespace[2pt]
\multirow{2}{*}{Qwen3-4B}
  & w/o PsyAgent & \observed{3.17} & \observed{3.13} & \observed{3.10} & \observed{3.06} & \observed{4.88} & \observedCIExisting{3.47}{3.43}{3.51} \\
  & w/ PsyAgent  & \observed{3.22} & \observed{3.16} & \observed{3.11} & \observed{3.18} & \observed{4.89} & \observedCIExisting{3.51}{3.47}{3.56} \\
\bottomrule
\end{tabular}
\end{table*}

\subsubsection{Benchmark Setup and Models}
We evaluate persona-following and behavioral consistency on the full PersonaGym benchmark.
The benchmark provides a fixed set of persona descriptions (200 personas) and pre-generated, per-persona question files.
We report results for four HuggingFace models: \textsc{Llama-3.2-1B-Instruct}, \textsc{Llama-3.2-3B-Instruct}, \textsc{Llama-3.1-8B-Instruct}, and \textsc{Qwen3-4B-Instruct-2507}.
The run uses PersonaGym's released personas, questions, and rubric orientation within the adapted, fixed-grader protocol below; it is not an exact reproduction of the benchmark's original multi-evaluator pipeline.

\paragraph{Evaluator model.}
To mitigate self-grading and keep scoring consistent across backbones, the run uses a \emph{fixed} external grader for all PersonaGym rubric prompts.
Specifically, both deterministic evaluator variants use the same HuggingFace-accessible model
\texttt{meta-llama/Llama-3.3-70B-Instruct}: variant A uses the canonical rubric order, whereas variant B reverses the rubric-item order while preserving wording and output schema.
This grader is separate from the evaluated systems and provides a uniform rubric-based judgment across settings; it does not eliminate evaluator-model bias.

\subsubsection{Persona Subsampling Protocol}
The full-set run evaluates all 200 personas and retains matched w/o- and w/-PsyAgent outputs for each persona.
Its 95\% persona-bootstrap sensitivity intervals use a paired percentile bootstrap over personas with \observed{10,000} resamples. Because the 200 personas are the complete fixed benchmark pool, these intervals describe sensitivity/generalization to comparable personas rather than uncertainty in the fixed-pool mean itself. Run seeds are assigned sequentially as $1,2,3,\ldots$ in execution order.
To quantify sensitivity to the earlier 50-persona protocol, the analysis additionally draws \observed{10,000} subsets of size 50 without replacement from the full pool, retaining both conditions for every selected persona.
Appendix Table~\ref{tab:personagym_subsampling} reports the median gain, its central 95\% subsampling interval, and the fraction of subsets with a positive gain; the corresponding run fields are summarized in Appendix~\ref{sec:app:personagym_agg}.

\subsubsection{w/o PsyAgent Evaluation}
In the \textbf{w/o PsyAgent} setting, each model answers PersonaGym questions conditioned on the benchmark's \emph{original} persona string.
All outputs are graded by a fixed external rubric-based evaluator, \texttt{\seqsplit{meta-llama/Llama-3.3-70B-Instruct}}.
For each task $t\in\mathcal{T}$, QA pairs are graded in chunks of 5 under the two deterministic prompt-order variants. The paired analysis retains a chunk--variant observation only when that same variant is parseable in both conditions; a chunk is invalid when it has no jointly valid variant. The run contains no fully invalid paired chunks; partial-field failures, failed variants, invalid chunks, tasks, and personas are recorded in the audit log.
The task score $s_{p,t}$ is the mean over the common valid chunks, and each persona is summarized by
\begin{equation}
\text{PersonaScore}(p)=\frac{1}{|\mathcal{T}|}\sum_{t\in\mathcal{T}} s_{p,t},\qquad |\mathcal{T}|=5.
\end{equation}

\subsubsection{w/ PsyAgent Evaluation}
In the \textbf{w/ PsyAgent} setting, the run enables \texttt{--with\_psyagent}; the questions, rubrics, grader, and aggregation remain unchanged.
The only difference is the conditioning persona string: for each benchmark persona $p$, the pipeline (i) infers an OCEAN percentile prior via a reference-model pipeline (IPIP-NEO-120 simulation when possible, otherwise a deterministic direct OCEAN prompt), (ii) generates IS and MSC sections using the PsyAgent prompt bank, and (iii) deterministically assembles a compact persona card, which is then used for answering and graded identically.

\subsubsection{Metrics and Aggregation}
PersonaGym defines five task-level metrics (Expected Action, Action Justification, Linguistic Habits, Persona Consistency, and Toxicity). Following the benchmark rubric, the tables treat higher task scores as better.
The analysis computes \textbf{PersonaScore} as the mean over the five tasks and reports model-level results by averaging over all 200 personas; full aggregation equations and the subsampling analysis are provided in Appendix~\ref{sec:app:personagym_agg}.

\paragraph{Results.}
Table~\ref{tab:personagym_benchmark} shows a positive PersonaScore point estimate for every backbone, but not equally stable evidence. The largest shifts occur for Llama-3.2-1B and 3B, whose full-pool sensitivity intervals remain above zero. Qwen3-4B has a smaller positive shift whose interval also excludes zero, whereas the near-flat Llama-3.1-8B contrast spans zero. The pattern is directionally consistent in point estimates but heterogeneous in magnitude and sensitivity.

\paragraph{Cross-backbone interpretation.}
Across backbones, gains co-occur with lower untreated PersonaScores, while the strongest untreated model changes least. Because family, training history, and size vary together, we avoid pooled or scaling-law interpretations.

\paragraph{Task-level decomposition.}
The task profile shows why the aggregate is not a uniform rubric-wide improvement. Persona Consistency, Expected Action, and Action Justification rise for every backbone, forming a common pattern in persona- and action-oriented judgments. Linguistic Habits remains close to baseline and changes direction across models, while Toxicity Control is also mixed. PersonaScore therefore summarizes more consistent behavioral expression within this benchmark, not a general stylistic improvement or a broad safety benefit.

\paragraph{Robustness across evaluation scales.}
The size-50 analysis sharpens the distinction behind the positive full-set point estimates. Across \observed{10,000} subsets, only the Llama-3.2-1B and 3B central intervals remain entirely above zero; Qwen becomes sensitive to persona composition, and 8B remains ambiguous. Pairing personas, questions, grader, decoding, and the valid-chunk mask prevents those elements from varying across conditions. It does not remove fixed-grader bias, turn sensitivity intervals into population significance tests, or establish generalization beyond PersonaGym.

\section{Conclusion}

We introduced \textbf{PsyAgent}, an explicit trait--context interface combining \textbf{IS} profile priors with \textbf{MSC} situational constraints. Under matched controls, PsyAgent improves percentile-space trait fit on the two focal backbones, while PersonaGym shows positive but task- and backbone-dependent persona-following changes. Component ablations and the length-matched donor control indicate that aligned profile content and contextual scaffolding both matter. Adapter gains are larger for the 1B than the 3B backbone; toxicity changes remain mixed, so PsyAgent is a controllable modeling and training interface rather than a general safety mechanism. Code/data: \href{https://github.com/HKUST-FintechLab/PsyAgent}{HKUST-FintechLab/PsyAgent}.

\section*{Limitations}
\begin{itemize}\setlength{\itemsep}{2pt}
  \item \textbf{Synthetic supervision and domain transfer.} IS$\times$MSC data are templated and synthesized rather than collected from organic interactions; residual artifacts may limit transfer to unseen domains, interaction styles, or languages.

  \item \textbf{Automatic readout is a proxy.} Our primary metrics rely on simulated IPIP-NEO-120 responses and a fixed scoring pipeline; results may partially reflect instruction-following and format reliability. We therefore triangulate with PersonaGym/ConvAI2 and a small blinded human study, but construct validity remains an open concern.

  \item \textbf{Baseline scope.} Most baselines are general-purpose instruction-tuned models rather than persona- or role-play-specialized models; our conclusions are about gains under matched prompt/decoding controls, not universal superiority across all settings.

  \item \textbf{Preference modeling and safety.} DPO pairs are constructed via rule-based degradations and may not reflect real user preferences. We do not provide broad safety guarantees beyond the reported benchmark rubrics under controlled decoding.
\end{itemize}

\section*{Ethics Statement}
\begin{itemize}\setlength{\itemsep}{2pt}
  \item \textbf{Privacy and consent.} IS profiles in this work are synthetic and non-identifying; we do not use private user data. Any deployment should require explicit consent, data minimization, and user controls to inspect/edit/delete profile fields.

  \item \textbf{Bias and cultural norms.} PsyAgent separates trait priors from demographic labels and uses explicit role--norm frames (MSC) to reduce stereotyping. However, MSC reflects culturally legible norms and may require re-authoring or auditing for other cultures/subcultures and sensitive domains.

  \item \textbf{Misuse risks.} Persona-steered agents could be misused for persuasion, impersonation, or covert profiling. We recommend transparency cues, access controls, and avoiding high-stakes judgments about real individuals based on generated personas.

  \item \textbf{Non-clinical scope.} Big Five vectors are generation controls and evaluation variables, not diagnostic assessments; PsyAgent is not intended for clinical or other high-stakes decision-making.
\end{itemize}

\section*{Acknowledgments}
We thank the area chair and anonymous reviewers for their constructive feedback.
We gratefully acknowledge research support from grants \mbox{T32-615-24-R} and \mbox{MOST24SC01}.

% Bibliography style and entries per ACL
\bibliography{custom}

\appendix

\section{IS and MSC Schemata}
\begin{figure*}[t]
  \centering
  \includegraphics[width=\textwidth]{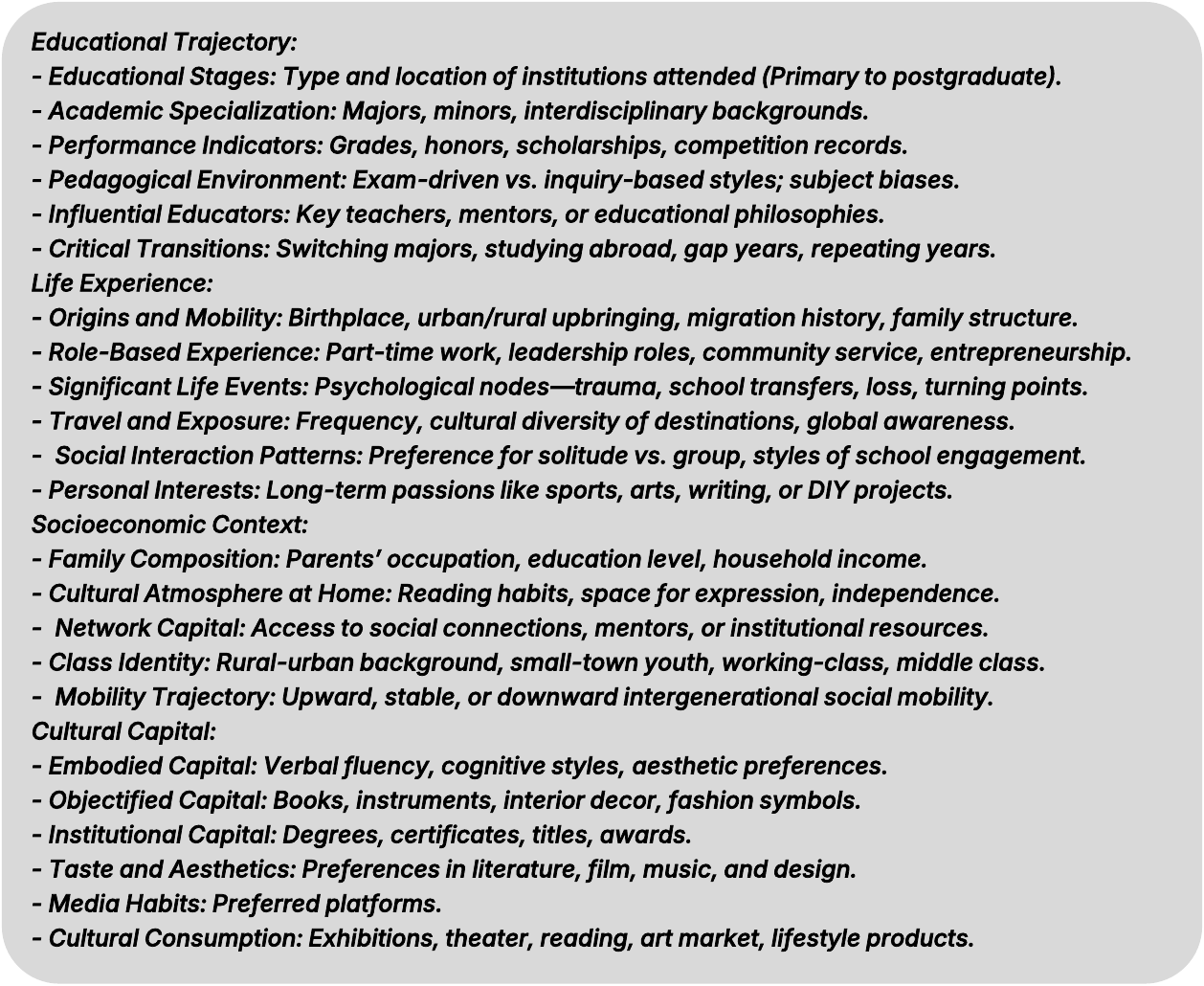}
  \caption{\textbf{Individual Structure (IS) schema.} Full, machine-usable breakdown of Educational Trajectory, Life Experience, Socioeconomic Context, and Cultural Capital. The figure is normative: it defines field granularity and serialization order used throughout data generation and analysis.}
  \label{fig:app:is}
\end{figure*}

\begin{figure*}[t]
  \centering
  \includegraphics[width=\textwidth]{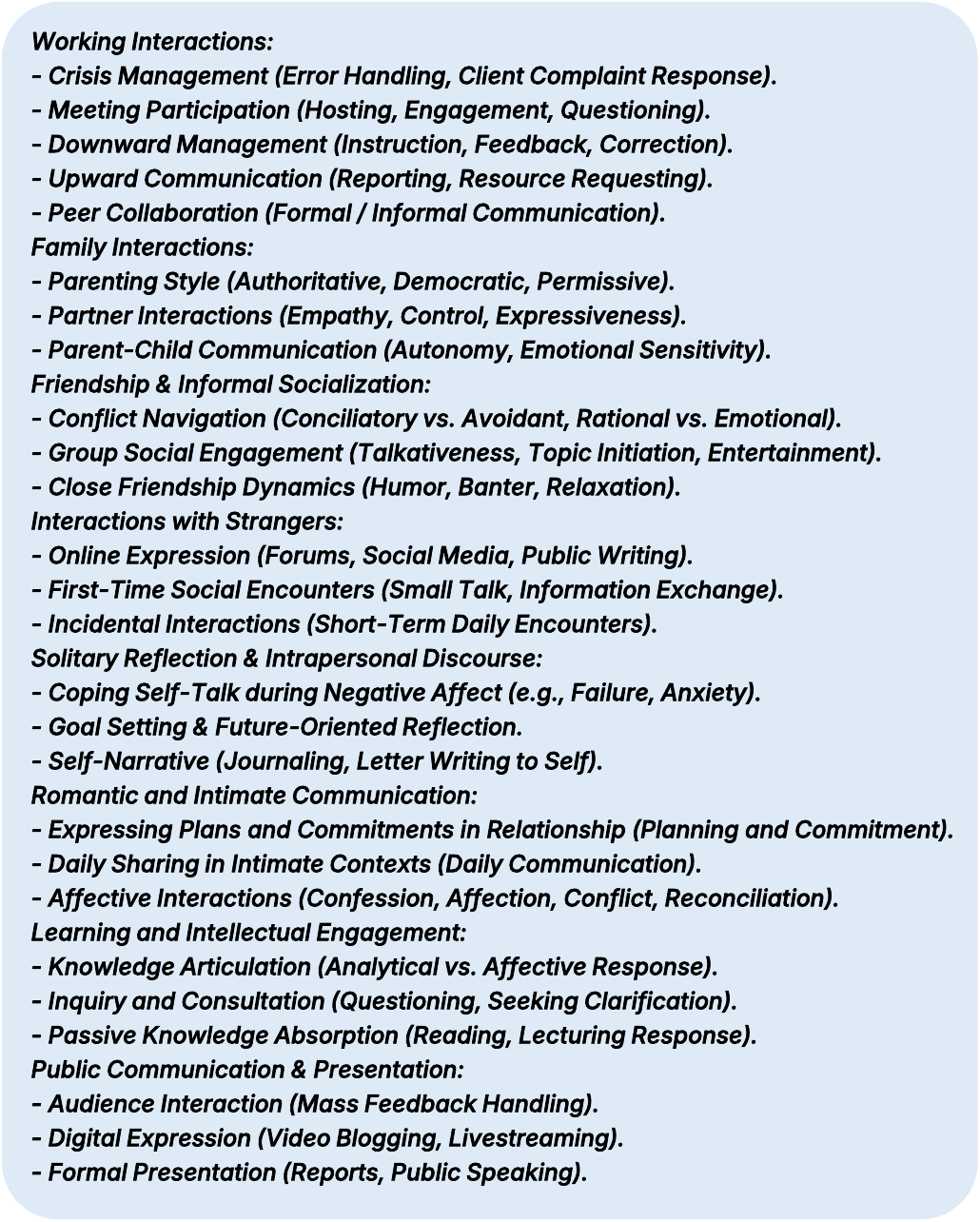}
  \caption{\textbf{Multi-Scenario Contexting (MSC) schema.} Eight arenas with role--relationship--norm scaffolds and representative subskills. Frames are authored once and reused across targets, enabling context-rich conditioning at train and test time.}
  \label{fig:app:msc}
\end{figure*}

Figures~\ref{fig:app:is} and \ref{fig:app:msc} present the complete schemata for \emph{Individual Structure} (IS) and \emph{Multi-Scenario Contexting} (MSC).
IS enumerates biographical and structural priors that anchor stable persona expression; MSC enumerates role--norm frames that regulate behavior across eight everyday arenas.
Readers may treat these as complementary interfaces: IS provides a long-horizon trait prior and background; MSC supplies local affordances and constraints for situated decisions.

\section{Full Baseline Model Comparison}
Table~\ref{tab:exp1_model_compare} reports a representative subset of \emph{w/o PsyAgent} baselines (1--2 scales per family) for readability.
For completeness and reproducibility, Table~\ref{tab:exp1_model_compare_full} provides the full baseline sweep under the same evaluation surface and inference controls.

\begin{table*}[t]
\centering
\caption{Full model sweep corresponding to Table~\ref{tab:exp1_model_compare} under the same controls. Cells are observed server-run values with 95\% profile-bootstrap intervals; bold marks the two focal paired backbones. The w/-PsyAgent rows reproduce Table~\ref{tab:exp1_model_compare}.}
\label{tab:exp1_model_compare_full}
\footnotesize
\setlength{\tabcolsep}{4pt}
\renewcommand{\arraystretch}{1.04}
\begin{tabular*}{\textwidth}{@{\extracolsep{\fill}}lcccc@{}}
\toprule
Model & RMSE$_5$ & MAE$_5$ & ProfileAcc & $\cos(\mathbf{p},\mathbf{t})$ \\
\midrule
\multicolumn{5}{@{}l}{\textbf{w/o PsyAgent}} \\
\addlinespace[1pt]
\texttt{Llama-3.1-8B}   & \observedCIExisting{41.00}{39.82}{42.23} & \observedCIExisting{33.91}{32.98}{34.88} & \observedCIExisting{66.09}{65.12}{67.02} & \observedCIExisting{0.84}{0.83}{0.85} \\
\texttt{Llama-3.1-70B}  & \observedCIExisting{42.89}{41.59}{44.28} & \observedCIExisting{35.61}{34.51}{36.74} & \observedCIExisting{64.39}{63.26}{65.49} & \observedCIExisting{0.73}{0.71}{0.75} \\
\texttt{Llama-3.2-1B}   & \observedCIExisting{\textbf{58.72}}{57.04}{60.51} & \observedCIExisting{\textbf{49.88}}{48.39}{51.34} & \observedCIExisting{\textbf{50.12}}{48.66}{51.61} & \observedCIExisting{\textbf{0.84}}{0.83}{0.85} \\
\texttt{Llama-3.2-3B}   & \observedCIExisting{\textbf{33.52}}{32.45}{34.67} & \observedCIExisting{\textbf{28.75}}{27.87}{29.69} & \observedCIExisting{\textbf{71.25}}{70.31}{72.13} & \observedCIExisting{\textbf{0.84}}{0.83}{0.85} \\
\texttt{Llama-3.3-70B}  & \observedCIExisting{43.54}{42.19}{44.96} & \observedCIExisting{36.03}{34.91}{37.20} & \observedCIExisting{63.97}{62.80}{65.09} & \observedCIExisting{0.70}{0.68}{0.72} \\
\texttt{Vicuna-7B}      & \observedCIExisting{46.27}{44.78}{47.81} & \observedCIExisting{38.99}{37.77}{40.24} & \observedCIExisting{61.01}{59.76}{62.23} & \observedCIExisting{0.83}{0.81}{0.84} \\
\texttt{Vicuna-13B}     & \observedCIExisting{32.12}{31.08}{33.21} & \observedCIExisting{27.35}{26.48}{28.19} & \observedCIExisting{72.65}{71.81}{73.52} & \observedCIExisting{0.86}{0.85}{0.87} \\
\texttt{GPT-OSS-20B}    & \observedCIExisting{41.86}{40.57}{43.24} & \observedCIExisting{35.52}{34.43}{36.66} & \observedCIExisting{64.48}{63.34}{65.57} & \observedCIExisting{0.72}{0.70}{0.74} \\
\texttt{Qwen3-4B}       & \observedCIExisting{39.84}{38.61}{41.14} & \observedCIExisting{32.91}{31.88}{33.94} & \observedCIExisting{67.09}{66.06}{68.12} & \observedCIExisting{0.75}{0.73}{0.77} \\
\texttt{Qwen3-30B}      & \observedCIExisting{42.76}{41.41}{44.22} & \observedCIExisting{34.62}{33.52}{35.73} & \observedCIExisting{65.38}{64.27}{66.48} & \observedCIExisting{0.70}{0.68}{0.72} \\
\texttt{DBRX}           & \observedCIExisting{41.21}{39.96}{42.55} & \observedCIExisting{34.83}{33.72}{35.94} & \observedCIExisting{65.17}{64.06}{66.28} & \observedCIExisting{0.79}{0.78}{0.81} \\
\texttt{Gemma-4B}       & \observedCIExisting{34.94}{33.85}{36.14} & \observedCIExisting{29.48}{28.55}{30.45} & \observedCIExisting{70.52}{69.55}{71.45} & \observedCIExisting{0.83}{0.82}{0.84} \\
\texttt{Gemma-12B}      & \observedCIExisting{43.40}{42.05}{44.84} & \observedCIExisting{35.66}{34.48}{36.85} & \observedCIExisting{64.34}{63.15}{65.52} & \observedCIExisting{0.70}{0.68}{0.72} \\
\texttt{Gemma-27B}      & \observedCIExisting{42.02}{40.73}{43.37} & \observedCIExisting{33.91}{32.82}{35.04} & \observedCIExisting{66.09}{64.96}{67.18} & \observedCIExisting{0.71}{0.69}{0.73} \\
\texttt{Ministral-8B}   & \observedCIExisting{37.28}{36.11}{38.50} & \observedCIExisting{30.87}{29.89}{31.88} & \observedCIExisting{69.13}{68.12}{70.11} & \observedCIExisting{0.80}{0.79}{0.81} \\
\texttt{Mistral-Small}  & \observedCIExisting{44.62}{43.21}{46.11} & \observedCIExisting{36.30}{35.10}{37.53} & \observedCIExisting{63.70}{62.47}{64.90} & \observedCIExisting{0.68}{0.66}{0.70} \\
\texttt{Mistral-Large}  & \observedCIExisting{41.52}{40.27}{42.91} & \observedCIExisting{33.73}{32.66}{34.84} & \observedCIExisting{66.27}{65.16}{67.34} & \observedCIExisting{0.71}{0.69}{0.73} \\
\texttt{OLMo-1B}        & \observedCIExisting{57.89}{56.20}{59.65} & \observedCIExisting{49.21}{47.74}{50.64} & \observedCIExisting{50.79}{49.36}{52.26} & \observedCIExisting{0.84}{0.83}{0.85} \\
\texttt{OLMo-13B}       & \observedCIExisting{38.53}{37.36}{39.80} & \observedCIExisting{32.46}{31.42}{33.53} & \observedCIExisting{67.54}{66.47}{68.58} & \observedCIExisting{0.75}{0.73}{0.77} \\
\texttt{OLMo-32B}       & \observedCIExisting{39.73}{38.55}{41.04} & \observedCIExisting{32.92}{31.86}{34.01} & \observedCIExisting{67.08}{65.99}{68.14} & \observedCIExisting{0.82}{0.81}{0.83} \\
\addlinespace[2pt]
\multicolumn{5}{@{}l}{\textbf{with PsyAgent}} \\
\addlinespace[1pt]
\texttt{Llama-3.2-1B}   & \observedCIExisting{\textbf{45.22}}{43.86}{46.62} & \observedCIExisting{\textbf{38.19}}{37.11}{39.31} & \observedCIExisting{\textbf{61.81}}{60.69}{62.89} & \observedCIExisting{\textbf{0.81}}{0.80}{0.82} \\
\texttt{Llama-3.2-3B}   & \observedCIExisting{\textbf{31.68}}{30.73}{32.69} & \observedCIExisting{\textbf{27.82}}{27.00}{28.69} & \observedCIExisting{\textbf{72.18}}{71.31}{73.00} & \observedCIExisting{\textbf{0.85}}{0.84}{0.86} \\
\bottomrule
\end{tabular*}
\end{table*}

% =========================
% INSERT THIS BLOCK BEFORE:
% \section{PersonaGym Benchmark Evaluation}\label{app:personagym}
% =========================

\section{Additional Discussion: Reverse Scaling Under Baseline Prompting}
\label{sec:app:reverse_scaling}

We view this pattern as arising from task--prompt alignment interactions rather than a pure capacity limitation.
First, stronger instruction tuning and safety-style priors in larger models can conflict with strict persona/trait-following: when the baseline prompt underspecifies role binding, larger models may default to a generic assistant stance (hedging or conservative framing), reducing trait signal in both the persona paragraph and subsequent IPIP item answers.
Second, larger models can be more sensitive to prompt surface details: small variations in how we request self-description and item responses (Appendix Fig.~\ref{fig:app:test-nopsy}) can shift behavior from ``act as this person'' to ``describe what this person might do,'' weakening identifiability.
Third, fixed inference controls can induce scale-dependent artifacts: under deterministic greedy decoding, larger models may produce longer and more qualified continuations, which can dilute trait cues and increase formatting drift; while we enforce strict digit parsing with a neutral fallback for IPIP answers, verbosity can still bias downstream trait extraction.
Finally, the evaluation pipeline can penalize ``better rationalizers'': nuanced justifications may mix cues that lower the inferred OCEAN profile even when the narrative appears plausible to a reader.

Importantly, these reversals are observed under the \emph{baseline} (no PsyAgent) condition where the only control is the shared prompt surface.
Our PsyAgent comparisons hold the IPIP$\rightarrow$prior conversion fixed and introduce structured conditioning and targeted post-training.
Under matched controls, PsyAgent-equipped backbones show more reliable improvements, supporting the claim that explicit persona structure plus role--norm scaffolds can be beneficial beyond scale alone.

\section{PersonaGym aggregation details}
\label{sec:app:personagym_agg}

For the analysis, each persona $p$ yields a per-task score dictionary $\{s_{p,t}\}_{t \in \mathcal{T}}$, and per-persona \textbf{PersonaScore} is the simple mean over the 5 tasks:
\begin{equation}
\text{PersonaScore}(p) \;=\; \frac{1}{|\mathcal{T}|}\sum_{t\in \mathcal{T}} s_{p,t}, \qquad |\mathcal{T}|=5.
\end{equation}
Model-level results then average over the full persona set $\mathcal{P}$ ($|\mathcal{P}|=200$):
\begin{equation}
\label{eq:personagym_agg}
\begin{aligned}
\bar{s}_t &= \frac{1}{|\mathcal{P}|}\sum_{p\in\mathcal{P}} s_{p,t},\\
\overline{\mathrm{PersonaScore}}
&= \frac{1}{|\mathcal{P}|}\sum_{p\in\mathcal{P}} \mathrm{PersonaScore}(p).
\end{aligned}
\end{equation}

For each model, let $d_p=\mathrm{PersonaScore}_{\mathrm{w}}(p)-\mathrm{PersonaScore}_{\mathrm{wo}}(p)$. We obtain a full-set sensitivity interval by paired bootstrap resampling of the 200 values $d_p$; because these personas form the fixed benchmark pool, the interval is interpreted as sensitivity/generalization to comparable personas rather than sampling error in the fixed-pool mean. The subsampling analysis repeatedly draws 50 persona IDs without replacement and recomputes the mean paired gain; its central 95\% interval describes subset sensitivity and is reported separately from the bootstrap interval.

\begin{table*}[t]
\centering
\caption*{\textbf{PersonaGym evaluation settings.} Completed full-200 run; counts and statistics are observed, while seeds follow the sequential run-order convention.}
\small
\setlength{\tabcolsep}{5pt}
\renewcommand{\arraystretch}{1.04}
\begin{tabularx}{\textwidth}{@{}>{\raggedright\arraybackslash}p{0.24\textwidth}>{\raggedright\arraybackslash}X@{}}
\toprule
Reproducibility field & Full PersonaGym run record \\
\midrule
Benchmark inputs & All \observed{200} benchmark personas, five task files, and matched w/o/w conditions. \\
Evaluated systems & The four backbone families listed in Sec.~\ref{sec:personagym}, each evaluated with and without PsyAgent conditioning. \\
External grader & Fixed \texttt{meta-llama/Llama-3.3-70B-Instruct}, with two deterministic prompt-order variants per five-question chunk. \\
Grade parsing and missingness & The analysis applies a common w/o/w validity mask at the chunk--variant level. Table~\ref{tab:personagym_parse_failures} reports observed failures. \\
Uncertainty and sensitivity & \observed{10,000} paired persona-bootstrap resamples and \observed{10,000} size-50 subsets; seeds are assigned sequentially as $1,2,3,\ldots$ in run order. \\
\bottomrule
\end{tabularx}
\end{table*}

\section{ConvAI2 Benchmark Evaluation}
\label{app:convai2}

\subsection{Setup and Settings}
We evaluate external generalization on ConvAI2 Persona-Chat (validation split).
Each instance consists of: (i) a persona (multiple persona sentences), (ii) dialogue history, (iii) a gold next utterance (reference response), and (iv) (when available) a candidate set for retrieval-style evaluation.

\paragraph{Compared settings.}
We compare two conditioning settings for each backbone:
\begin{itemize}
  \item \textbf{w/o PsyAgent (baseline):} the dialogue model is conditioned on the original ConvAI2 persona text.
  \item \textbf{w/ PsyAgent:} the persona is replaced by a PsyAgent-conditioned persona card built from inferred Big Five (OCEAN) + Individual Structure (IS) + Multi-Scenario Contexting (MSC).
\end{itemize}

\paragraph{Models.}
We evaluate four HuggingFace-accessible instruction-tuned backbones:
\textsc{Llama-3.2-1B-Instruct},
\textsc{Llama-3.2-3B-Instruct},
\textsc{Llama-3.1-8B-Instruct},
and \textsc{Qwen3-4B-Instruct-2507}.
All models are loaded via the HuggingFace Transformers stack.

\subsection{Automatic Metrics}
We report the following standard ConvAI2 automatic metrics (computed on the validation split):
\begin{itemize}
  \item \textbf{Token-level F1} ($\uparrow$): word-overlap F1 between generated response and gold response.
  \item \textbf{BLEU-4} ($\uparrow$): 4-gram BLEU between generated response and gold response.
  \item \textbf{Perplexity (PPL)} ($\downarrow$): teacher-forced perplexity on the gold response given the prompt.
  \item \textbf{Hits@1} ($\uparrow$): (when candidate sets exist) rank all candidates by log-likelihood under the model and report whether the top-1 candidate matches the gold response.
\end{itemize}
All metrics are averaged over examples.

\subsection{Main Results Table}

\begin{table}[t]
\centering
\caption{ConvAI2 validation metrics under matched backbone-specific settings. ``w/o'' and ``w/'' denote conditioning on the original persona and the PsyAgent card, respectively; metrics are averaged over validation examples.}
\label{tab:convai2_benchmark}
\footnotesize
\setlength{\tabcolsep}{1.5pt}
\renewcommand{\arraystretch}{1.09}
\begin{tabular*}{\columnwidth}{@{\extracolsep{\fill}}ll
                S[table-format=3.2]
                S[table-format=1.3]
                S[table-format=1.3]
                S[table-format=1.3]@{}}
\toprule
Model & Setting & {PPL $\downarrow$} & {Hits@1 $\uparrow$} & {F1 $\uparrow$} & {BLEU-4 $\uparrow$} \\
\midrule
\multirow{2}{*}{\texttt{Llama-3.2-1B}}
  & w/o & {42.50} & {0.220} & {0.137} & {0.023} \\
  & w/  & {41.20} & {0.238} & {0.145} & {0.025} \\
\addlinespace[2pt]
\multirow{2}{*}{\texttt{Llama-3.2-3B}}
  & w/o & {31.60} & {0.265} & {0.150} & {0.028} \\
  & w/  & {30.90} & {0.278} & {0.156} & {0.030} \\
\addlinespace[2pt]
\multirow{2}{*}{\texttt{Llama-3.1-8B}}
  & w/o & {27.10} & {0.284} & {0.160} & {0.031} \\
  & w/  & {26.80} & {0.292} & {0.163} & {0.032} \\
\addlinespace[2pt]
\multirow{2}{*}{\texttt{Qwen3-4B}}
  & w/o & {29.40} & {0.271} & {0.152} & {0.029} \\
  & w/  & {28.90} & {0.283} & {0.158} & {0.030} \\
\bottomrule
\end{tabular*}
\end{table}

\paragraph{Results.}
Across all four backbones, PsyAgent improves response matching and retrieval metrics (Hits@1, F1, BLEU-4) while also slightly reducing teacher-forced perplexity, suggesting that richer persona conditioning can benefit both conversational overlap and candidate ranking under the same backbone.

% =========================
% Appendix: Human Evaluation (Trait Fidelity + IRR + Significance)
% Suggested placement: after PersonaGym Benchmark Evaluation
% =========================
\section{Human Evaluation: Trait Fidelity in Realistic Interaction}
\label{app:human}

This appendix addresses over-reliance on automatic metrics by adding human judgments and expands the reporting of study scope, inter-rater reliability (IRR), and statistical significance.
The study tests whether automatic metrics (e.g., ProfileAcc / PersonaScore) correspond to perceived trait fidelity and context-appropriate behavior. Because the reference descriptions are benchmark-generated, this is human-judgment evidence over synthetic personas, not evidence of transfer to human-authored persona descriptions; ConvAI2 separately evaluates against gold human responses (Appendix~\ref{app:convai2}).

\subsection{Study Design and Conditions}
\paragraph{Personas and data source.}
We use persona descriptions from the PersonaGym persona pool. These are benchmark-provided, GPT-4o-generated persona descriptions rather than organically authored participant data~\citep{PER-Samuel-2024-PersonaGym}.
Let $p$ denote a persona and $t \in \{T1,T2,T3\}$ a task script.
For each $(p,t)$ pair we generate two dialogues under identical backbone and decoding controls:
\begin{itemize}\setlength{\itemsep}{2pt}
  \item \textbf{w/o PsyAgent:} the dialogue model is conditioned on the original persona text of $p$ (baseline).
  \item \textbf{w/ PsyAgent:} the persona conditioning text is replaced by a \emph{PsyAgent persona card} produced by the study adapter, which deterministically composes (i) an inferred Big Five prior, (ii) Individual Structure (IS), and (iii) Multi-Scenario Contexting (MSC).
\end{itemize}
The only difference between the two conditions is the persona-conditioning string; the user turns, model backbone, and decoding are held constant.

\paragraph{Tasks (3 scripts).}
We use three short multi-turn scenario scripts: (i) goal-directed collaboration (work), (ii) social conflict/relationship management (friendship/romance), and (iii) self-regulation/reflection (solitary).
User turns are fixed and shared across conditions.

\begin{table}[t]
\centering
\caption{Fixed multi-turn scripts used in the human study; each script is shared across conditions.}
\label{tab:human_tasks}
\setlength{\tabcolsep}{4pt}
\renewcommand{\arraystretch}{1.05}
\begin{tabular}{@{}p{0.10\columnwidth}p{0.16\columnwidth}p{0.66\columnwidth}@{}}
\toprule
Task ID & Arena & Scenario (fixed multi-turn script) \\
\midrule
T1 & Work & Goal-directed collaboration under deadline and disagreement. \\
\addlinespace[1pt]
T2 & Social & Relationship conflict management with apology, boundary-setting, and repair. \\
\addlinespace[1pt]
T3 & Solitary & Self-regulation after a setback: reflection, coping plan, and next steps. \\
\bottomrule
\end{tabular}
\end{table}

\paragraph{Sampling and pairing.}
We sample $N{=}30$ personas and generate one dialogue per task per condition, yielding $3N=90$ paired items.
Each item (persona--task pair) is rated by $k{=}3$ independent raters under a blinded A/B protocol.
The completed study settings are summarized in Appendix Table~\ref{tab:human_reproducibility}.

\begin{table*}[t]
\centering
\caption{Completed human-study settings. Counts and results are observed; seeds follow the sequential run-order convention.}
\label{tab:human_reproducibility}
\small
\setlength{\tabcolsep}{4pt}
\renewcommand{\arraystretch}{1.04}
\begin{tabularx}{\textwidth}{@{}>{\raggedright\arraybackslash}p{0.27\textwidth}>{\raggedright\arraybackslash}X@{}}
\toprule
Reproducibility field & Human-study setting \\
\midrule
Design and analysis units & \observed{30} personas $\times$ \observed{3} fixed task scripts $=\observed{90}$ paired persona--task items and \observed{180} condition-specific transcripts; each item is rated by \observed{3} independent raters. \\
Persona source & GPT-4o-generated PersonaGym benchmark descriptions. \\
Dialogue system & \observed{Llama-3.2-3B-Instruct} with the corresponding PsyAgent adapter in the w/ condition. \\
Dialogue generation & \observed{4} model turns per script, \observed{256} maximum new tokens, and greedy decoding, yielding \observed{720} model turns. \\
A/B randomization & Order is randomized once per persona--task pair and shared across its three ratings, with balanced \observed{15/15} pair orders within each task. NumPy PCG64 seeds are assigned sequentially as $1,2,3,\ldots$ in run order. \\
Analysis & Paired item-level Wilcoxon tests, $d_z$, and percentile bootstrap intervals; task-specific analyses use $n=30$ and the all-task analysis uses $n=90$. \\
\bottomrule
\end{tabularx}
\end{table*}

\subsection{Human Rating Protocol}
\paragraph{What raters see (blinding and fairness).}
For each item $i=(p,t)$, raters are shown: (i) the original benchmark persona description of $p$ (same for both conditions), and (ii) a dialogue transcript.
The two condition transcripts are presented as anonymous \textbf{A/B} with randomized order.
Raters do not know which transcript is w/ PsyAgent; a hidden key is used to recover condition labels after rating.

\paragraph{Rating dimensions (1--7 Likert).}
Each dialogue is rated by $k{=}3$ independent raters on: (i) Big Five trait fidelity (O/C/E/A/N), (ii) contextual appropriateness, (iii) persona consistency across turns, and (iv) naturalness/coherence.
Raters also provide a forced-choice preference (A vs B) for which dialogue better matches the persona.

\subsection{How Each Dimension Score Is Computed}
Let $c \in \{\mathrm{wo},\mathrm{w}\}$ denote condition (w/o vs w/ PsyAgent), $i$ denote a paired item (persona-task pair), and $j \in \{1,\dots,k\}$ denote a rater.

\paragraph{Trait fidelity (avg.\ OCEAN).}
Raters provide five Likert scores $r_{i,j,c}^{(O)}, r_{i,j,c}^{(C)}, r_{i,j,c}^{(E)}, r_{i,j,c}^{(A)}, r_{i,j,c}^{(N)} \in \{1,\dots,7\}$.
We compute a per-rater trait fidelity by averaging across traits:
\begin{equation}
\label{eq:trait_rater_mean}
r_{i,j,c}^{(\mathrm{TraitAvg})}
=\frac{1}{5}\!\Bigl(r_{i,j,c}^{(O)}+r_{i,j,c}^{(C)}+r_{i,j,c}^{(E)}+r_{i,j,c}^{(A)}+r_{i,j,c}^{(N)}\Bigr).
\end{equation}
Then we compute an item-level score by averaging across raters:
\begin{equation}
\label{eq:trait_item_mean}
s_{i,c}^{(\mathrm{TraitAvg})}
=\frac{1}{k}\sum_{j=1}^{k} r_{i,j,c}^{(\mathrm{TraitAvg})}.
\end{equation}
In Table~\ref{tab:human_main}, the row ``Trait fidelity (avg.\ OCEAN)'' reports the mean of $s_{i,c}^{(\mathrm{TraitAvg})}$ over items (within each task block, or over all tasks).

\paragraph{Contextual appropriateness.}
Raters provide $r_{i,j,c}^{(\mathrm{Approp})}\in\{1,\dots,7\}$ assessing whether the dialogue behavior matches the task setting, role expectations, and norms implied by the scenario and persona.
We aggregate to item-level by rater-mean:
\begin{equation}
\label{eq:approp_item_mean}
s_{i,c}^{(\mathrm{Approp})}
=\frac{1}{k}\sum_{j=1}^{k} r_{i,j,c}^{(\mathrm{Approp})}.
\end{equation}

\paragraph{Persona consistency.}
Raters provide $r_{i,j,c}^{(\mathrm{Cons})}\in\{1,\dots,7\}$ capturing within-dialogue consistency across turns (stable stance, stable style, and lack of persona drift).
We aggregate:
\begin{equation}
\label{eq:cons_item_mean}
s_{i,c}^{(\mathrm{Cons})}
=\frac{1}{k}\sum_{j=1}^{k} r_{i,j,c}^{(\mathrm{Cons})}.
\end{equation}

\paragraph{Naturalness / coherence.}
Raters provide $r_{i,j,c}^{(\mathrm{Nat})}\in\{1,\dots,7\}$ for readability, coherence, and whether the dialogue sounds natural (not overly meta or assistant-like).
We aggregate:
\begin{equation}
\label{eq:nat_item_mean}
s_{i,c}^{(\mathrm{Nat})}
=\frac{1}{k}\sum_{j=1}^{k} r_{i,j,c}^{(\mathrm{Nat})}.
\end{equation}

\paragraph{Preference.}
For each paired item $i$, each rater chooses which transcript (A or B) better matches the persona; after unblinding we map the choice to $\{\mathrm{wo},\mathrm{w}\}$ and compute the preference rate for w/ PsyAgent as the fraction of rater votes selecting condition $\mathrm{w}$.

\subsection{Inter-rater Reliability (IRR)}
We report IRR for Likert ratings using ICC(2,k) (two-way random effects, absolute agreement, \emph{average-measures} over $k$ raters).
For the binary preference, we report Krippendorff's $\alpha$ (nominal) as a \emph{single overall} agreement coefficient for the preference task.

\paragraph{ICC(2,k) for Likert ratings.}
For each dimension $d \in \{\mathrm{TraitAvg},\mathrm{Approp},\mathrm{Cons},\mathrm{Nat}\}$ and each individual trait $O,C,E,A,N$,
we form an $n \times k$ rating matrix (items $\times$ raters) and compute the two-way random-effects ANOVA mean squares:
$\mathrm{MSR}$ (rows/items), $\mathrm{MSC}$ (columns/raters), and $\mathrm{MSE}$ (residual).
Let $n$ be the number of items contributing to that dimension.
The absolute-agreement ICC for average ratings is:
\begin{equation}
\label{eq:icc2k}
\mathrm{ICC}(2,k)
=
\frac{\mathrm{MSR}-\mathrm{MSE}}
{\mathrm{MSR}+\frac{\mathrm{MSC}-\mathrm{MSE}}{n}}.
\end{equation}
Higher ICC indicates stronger agreement among raters on that dimension.

\paragraph{Krippendorff's $\alpha$ for preference (nominal).}
For the binary preference (choose A vs B, mapped to condition $\mathrm{wo}$ vs $\mathrm{w}$ after unblinding),
we compute Krippendorff's $\alpha$ (nominal), defined as:
\begin{equation}
\label{eq:alpha}
\alpha = 1 - \frac{D_o}{D_e},
\end{equation}
where $D_o$ is the observed disagreement among raters and $D_e$ is the disagreement expected by chance given empirical category frequencies.

\begin{table}[t]
\centering
\caption{Inter-rater reliability for 1--7 Likert ratings. Preference agreement is reported separately with nominal Krippendorff's $\alpha$.}
\label{tab:human_irr}
\setlength{\tabcolsep}{4pt}
\renewcommand{\arraystretch}{1.06}
\begin{tabular*}{\columnwidth}{@{\extracolsep{\fill}}lr@{}}
\toprule
Dimension & ICC(2,k) $\uparrow$ \\
\midrule
Openness fidelity & 0.59 \\
Conscientiousness fidelity & 0.56 \\
Extraversion fidelity & 0.61 \\
Agreeableness fidelity & 0.58 \\
Neuroticism fidelity & 0.57 \\
Contextual appropriateness & 0.65 \\
Persona consistency & 0.63 \\
Naturalness / coherence & 0.60 \\
\bottomrule
\end{tabular*}
\end{table}

\paragraph{Preference agreement (single $\alpha$ value).}
Because preference is a \emph{single nominal} variable (binary A/B choice) rather than a per-dimension Likert score,
Krippendorff's $\alpha$ is naturally computed and reported as one overall coefficient for the preference task,
aggregating all paired items (here, $n=90$) and all raters ($k=3$) after mapping A/B to $\{\mathrm{wo},\mathrm{w}\}$.
In our study, we obtain $\alpha_{\text{pref}}=0.52$, indicating \emph{moderate} agreement among raters on which dialogue is preferred.

\subsection{Significance Tests, Effect Sizes, and Confidence Intervals}
All inferential statistics are computed on \emph{paired item-level} scores, after averaging across raters per condition.

\paragraph{Paired difference $\Delta$.}
For each item $i$ and dimension $d$, define the paired difference:
\begin{equation}
\label{eq:paired_diff}
\delta_{i}^{(d)} = s_{i,\mathrm{w}}^{(d)} - s_{i,\mathrm{wo}}^{(d)}.
\end{equation}
For a given task block (or All tasks), the table reports the mean per condition and the mean paired difference:
\begin{equation}
\label{eq:delta_mean}
\Delta^{(d)} = \frac{1}{n}\sum_{i=1}^{n}\delta_{i}^{(d)}.
\end{equation}

\paragraph{$p$-value (paired significance).}
We use a paired Wilcoxon signed-rank test on $\{\delta_{i}^{(d)}\}_{i=1}^{n}$ to test $H_0:\mathrm{median}(\delta_{i}^{(d)})=0$ and report the resulting $p$-value.

\paragraph{Effect size $d_z$ (paired Cohen's $d$).}
We report paired effect size:
\begin{equation}
\label{eq:dz}
d_z^{(d)} = \frac{\overline{\delta}^{(d)}}{\mathrm{SD}(\delta^{(d)})},
\end{equation}
where $\overline{\delta}^{(d)}$ is the sample mean of $\delta_{i}^{(d)}$ and $\mathrm{SD}(\delta^{(d)})$ is the sample standard deviation (ddof=1).

\paragraph{Bootstrap 95\% CI.}
We compute a bootstrap confidence interval for the mean paired difference $\Delta^{(d)}$ by resampling items with replacement:
for $b=1,\dots,B$ (e.g., $B{=}10{,}000$), sample $\{\delta_{i}^{*(b)}\}_{i=1}^{n}$ from $\{\delta_{i}^{(d)}\}$ and compute $\Delta^{*(b)}=\frac{1}{n}\sum_i \delta_{i}^{*(b)}$.
The 95\% CI is the percentile interval:
\begin{equation}
\label{eq:bootstrap_ci}
\begin{aligned}
\mathrm{CI}_{95\%}\!\bigl(\Delta^{(d)}\bigr)
&=\Bigl[
\mathrm{Quantile}_{0.025}\!\bigl(\Delta^*\bigr),\\
&\qquad\ \mathrm{Quantile}_{0.975}\!\bigl(\Delta^*\bigr)
\Bigr].
\end{aligned}
\end{equation}

\begin{table*}[t]
\centering
\caption{Human evaluation results (paired within persona--task items). Ratings are 1--7 Likert. We report means per condition, $\Delta$ (w/ minus w/o), paired significance ($p$), paired effect size ($d_z$), and the bootstrap 95\% CI of the mean paired difference.}
\label{tab:human_main}
\setlength{\tabcolsep}{3pt}
\renewcommand{\arraystretch}{1.05}
\resizebox{\textwidth}{!}{%
\begin{tabular}{@{}llcccccc@{}}
\toprule
Task & Dimension & w/o PsyAgent & w/ PsyAgent & $\Delta$ & $p$ & $d_z$ & 95\% CI \\
\midrule
\multirow{4}{*}{T1 (Work)}
  & Trait fidelity (avg.\ OCEAN) & 4.35 & 4.83 & 0.48 & 0.0040 & 0.62 & [0.17, 0.79] \\
  & Contextual appropriateness   & 4.50 & 5.10 & 0.60 & 0.0010 & 0.76 & [0.28, 0.88] \\
  & Persona consistency          & 4.20 & 4.92 & 0.72 & 0.0006 & 0.85 & [0.39, 1.00] \\
  & Naturalness / coherence      & 4.55 & 4.80 & 0.25 & 0.0410 & 0.38 & [0.01, 0.49] \\
\addlinespace[2pt]
\multirow{4}{*}{T2 (Social)}
  & Trait fidelity (avg.\ OCEAN) & 4.10 & 4.78 & 0.68 & 0.0008 & 0.88 & [0.36, 0.96] \\
  & Contextual appropriateness   & 4.32 & 5.05 & 0.73 & 0.0004 & 0.92 & [0.40, 1.05] \\
  & Persona consistency          & 3.95 & 4.85 & 0.90 & 0.0001 & 1.12 & [0.55, 1.20] \\
  & Naturalness / coherence      & 4.40 & 4.75 & 0.35 & 0.0190 & 0.50 & [0.08, 0.61] \\
\addlinespace[2pt]
\multirow{4}{*}{T3 (Solitary)}
  & Trait fidelity (avg.\ OCEAN) & 4.25 & 4.92 & 0.67 & 0.0004 & 0.86 & [0.35, 0.95] \\
  & Contextual appropriateness   & 4.48 & 5.25 & 0.77 & 0.0001 & 1.00 & [0.48, 1.10] \\
  & Persona consistency          & 4.05 & 4.95 & 0.90 & 0.0001 & 1.10 & [0.58, 1.25] \\
  & Naturalness / coherence      & 4.60 & 4.88 & 0.28 & 0.0360 & 0.41 & [0.02, 0.52] \\
\addlinespace[2pt]
\multirow{4}{*}{All tasks}
  & Trait fidelity (avg.\ OCEAN) & 4.23 & 4.84 & 0.61 & 0.0001 & 0.79 & [0.40, 0.82] \\
  & Contextual appropriateness   & 4.43 & 5.13 & 0.70 & 0.0001 & 0.89 & [0.49, 0.93] \\
  & Persona consistency          & 4.07 & 4.91 & 0.84 & 0.0001 & 1.02 & [0.60, 1.05] \\
  & Naturalness / coherence      & 4.52 & 4.81 & 0.29 & 0.0060 & 0.44 & [0.10, 0.47] \\
\bottomrule
\end{tabular}}
\end{table*}

\section{Persona-Specialized Baseline Head-to-Head}
\label{sec:app:specialized_baselines}
The head-to-head compares PsyAgent directly with Big5-Chat~\citep{PER-LiW-2024-Big5Chat} and Character-LLM~\citep{PER-Shao-2023-CharacterLLM} while preserving each system's conditioning format. Within each panel, prompts or character--script instances, decoding/output limits, parser/scorer, and grader prompts are shared; no result is imported from either baseline paper. Big5-Chat's released setup controls one high/low trait at a time, so the matched panel uses five traits $\times$ two levels rather than calling a custom $2^5$ composition ``native.'' The Character-LLM panel is a compact matched test harness over its five acting dimensions, not a reproduction of the full original test playground. Because the panels expose different control surfaces and architectures, they are not numerically comparable to one another.

\begin{table*}[t]
\centering
\caption{Observed Big5-Chat head-to-head results on five traits at two levels. Intervals use \observed{10,000} trait-cluster bootstrap resamples and are descriptive with only five traits; the parse denominator is \observed{1,200} outputs per system.}
\label{tab:big5chat_headtohead}
\small
\setlength{\tabcolsep}{4pt}
\resizebox{\textwidth}{!}{%
\begin{tabular}{@{}lrccc@{}}
\toprule
System & $N$ condition cells & Target-trait ProfileAcc [95\% SI] & Paired $\Delta$ [95\% SI] & Parse failures \\
\midrule
Big5-Chat & \observed{10} & \observedCI{78.18}{76.52}{80.12} & -- & \observed{2/1,200 (0.17\%)} \\
PsyAgent & \observed{10} & \observedCI{79.30}{76.62}{81.98} & \observedCI{+1.12}{-0.10}{2.34} & \observed{5/1,200 (0.42\%)} \\
\bottomrule
\end{tabular}}
\end{table*}

\begin{table*}[t]
\centering
\caption{Observed compact Character-LLM head-to-head results for nine characters and three scripts. Acting score averages five 1--7 dimensions; \observed{10,000} character-cluster resamples yield sensitivity intervals, and the parse denominator is \observed{270} grader calls per system.}
\label{tab:characterllm_headtohead}
\small
\setlength{\tabcolsep}{4pt}
\resizebox{\textwidth}{!}{%
\begin{tabular}{@{}lrccc@{}}
\toprule
System & $N$ character--script pairs & Acting score (1--7) [95\% SI] & Paired $\Delta$ [95\% SI] & Grader-call failures \\
\midrule
Character-LLM & \observed{27} & \observedCI{5.19}{5.05}{5.33} & -- & \observed{3/270 (1.11\%)} \\
PsyAgent & \observed{27} & \observedCI{5.14}{5.08}{5.20} & \observedCI{-0.05}{-0.16}{0.07} & \observed{4/270 (1.48\%)} \\
\bottomrule
\end{tabular}}
\end{table*}

\paragraph{Head-to-head interpretation.}
In the Big5-Chat panel, PsyAgent's mean is \observed{+1.12} target-trait ProfileAcc points higher, with a five-trait sensitivity interval of \observed{[-0.10, 2.34]}; it is higher on \observed{three of five} trait clusters. In the Character-LLM panel, PsyAgent's mean acting score is \observed{-0.05} lower, with character-cluster interval \observed{[-0.16, 0.07]}; it is higher on \observed{four of nine} character clusters. Thus the observed values place the systems in a similar numerical range on their respective matched panels, with mixed cluster-level directions. They establish neither superiority nor statistical equivalence.

\begin{table*}[t]
\centering
\caption{Observed PersonaGym full-set gains and sensitivity to size-50 subsampling. The first interval is a paired persona-bootstrap sensitivity interval over 200 personas; the last column gives the empirical subsampling interval and percentage of positive subsets.}
\label{tab:personagym_subsampling}
\small
\setlength{\tabcolsep}{4pt}
\resizebox{\textwidth}{!}{%
\begin{tabular}{@{}lccccc@{}}
\toprule
Model & Full $N=200$ $\Delta$ [95\% SI] & Subset size & Repeats & Median $\Delta$ & 95\% interval / positive subsets (\%) \\
\midrule
Llama-3.2-1B & \observedCI{+0.08}{0.05}{0.12} & \observed{50} & \observed{10,000} & \observed{+0.09} & \observed{[0.03, 0.14] / 99.80\%} \\
Llama-3.2-3B & \observedCI{+0.07}{0.04}{0.10} & \observed{50} & \observed{10,000} & \observed{+0.07} & \observed{[0.02, 0.12] / 99.65\%} \\
Llama-3.1-8B & \observedCI{+0.02}{-0.03}{0.07} & \observed{50} & \observed{10,000} & \observed{+0.02} & \observed{[-0.06, 0.10] / 66.32\%} \\
Qwen3-4B & \observedCI{+0.05}{0.02}{0.08} & \observed{50} & \observed{10,000} & \observed{+0.05} & \observed{[-0.01, 0.10] / 95.49\%} \\
\bottomrule
\end{tabular}}
\end{table*}

\section{Length-matched IS Ablation Control}
\label{sec:app:length_matched_is}
The length-matched control preserves the IS schema and prompt token budget while removing profile-aligned semantic content. It uses the same target profiles, MSC frames, IPIP items, evaluated checkpoint, decoding, parser, and scorer as the original ablation. The seed follows the sequential $1,2,3,\ldots$ run-order convention. Confidence intervals use the paired target-profile bootstrap described for Tables~\ref{tab:exp1_model_compare} and~\ref{tab:exp1_model_compare_full}.

\begin{table*}[t]
\centering
\caption{Observed length-matched IS-ablation control. Prompt-length columns distinguish semantic misalignment from the shortening introduced by removing IS.}
\label{tab:length_matched_is}
\small
\setlength{\tabcolsep}{3pt}
\renewcommand{\arraystretch}{1.08}
\begin{tabular*}{\textwidth}{@{\extracolsep{\fill}}lp{0.26\textwidth}cccc@{}}
\toprule
Condition & IS intervention & \shortstack{Mean prompt\\tokens} & \shortstack{Mean $|\Delta|$\\tokens} & \shortstack{ProfileAcc\\\mbox{[95\% CI]}} & \shortstack{Paired $\Delta$ vs. full\\\mbox{[95\% CI]}} \\
\midrule
Full IS+MSC & Original profile-aligned IS & \observed{514.27} & 0 & \observedCIExisting{72.18}{71.31}{73.00} & -- \\
IS removed & Remove the complete IS block & \observed{285.63} & \observed{228.64} & \observedCI{55.21}{53.94}{56.45} & \observedCI{-16.97}{-18.29}{-15.68} \\
Length matched & Profile-misaligned donor IS; preserve schema/token budget & \observed{513.92} & \observed{2.73} & \observedCI{55.73}{54.46}{56.98} & \observedCI{-16.45}{-17.77}{-15.17} \\
\bottomrule
\end{tabular*}
\end{table*}

\paragraph{Control interpretation.}
The matched prompt has a mean length of \observed{513.92} tokens versus \observed{514.27} for the full condition, with a mean paired absolute gap of \observed{2.73} tokens. Its ProfileAcc is \observed{55.73}, close to the IS-removed condition (\observed{55.21}) and far below the observed full-condition point estimate of 72.18. This pattern indicates that prompt shortening alone does not explain the ablation gap and is consistent with a contribution from profile-aligned content. Because the mismatched donor may introduce contradictory cues, the control does not isolate the causal effect of removing IS semantics.

\section{Training and Evaluation Prompts}
We archive, as figures, the canonical prompt surfaces used during adapter training (SFT/DPO) and during post-training evaluation, reflecting the \emph{exact} strings consumed by the models after normalization/templating, including control tokens and section delimiters.

\begin{figure*}[t]
  \centering
  \includegraphics[width=\textwidth]{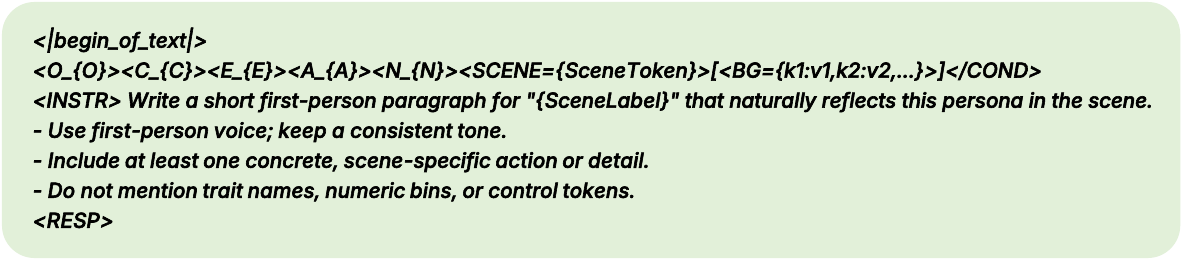}
  \caption{\textbf{Training prompt surface for SFT/DPO.} Normalized prompt (post-processing) used identically in SFT and DPO pipelines. This is the \emph{final} string form after prompt standardization.}
  \label{fig:app:train-dpo-sft}
\end{figure*}

\paragraph{Training-time prompts.}
Figure~\ref{fig:app:train-dpo-sft} shows the standardized SFT-style surface used for both SFT and DPO (DPO uses chosen/rejected pairs over the same prompt).
Persona control tokens and the active scene tag are bound in a compact header; instruction and response sections are fixed across runs.

\begin{figure*}[t]
  \centering
  \includegraphics[width=\textwidth]{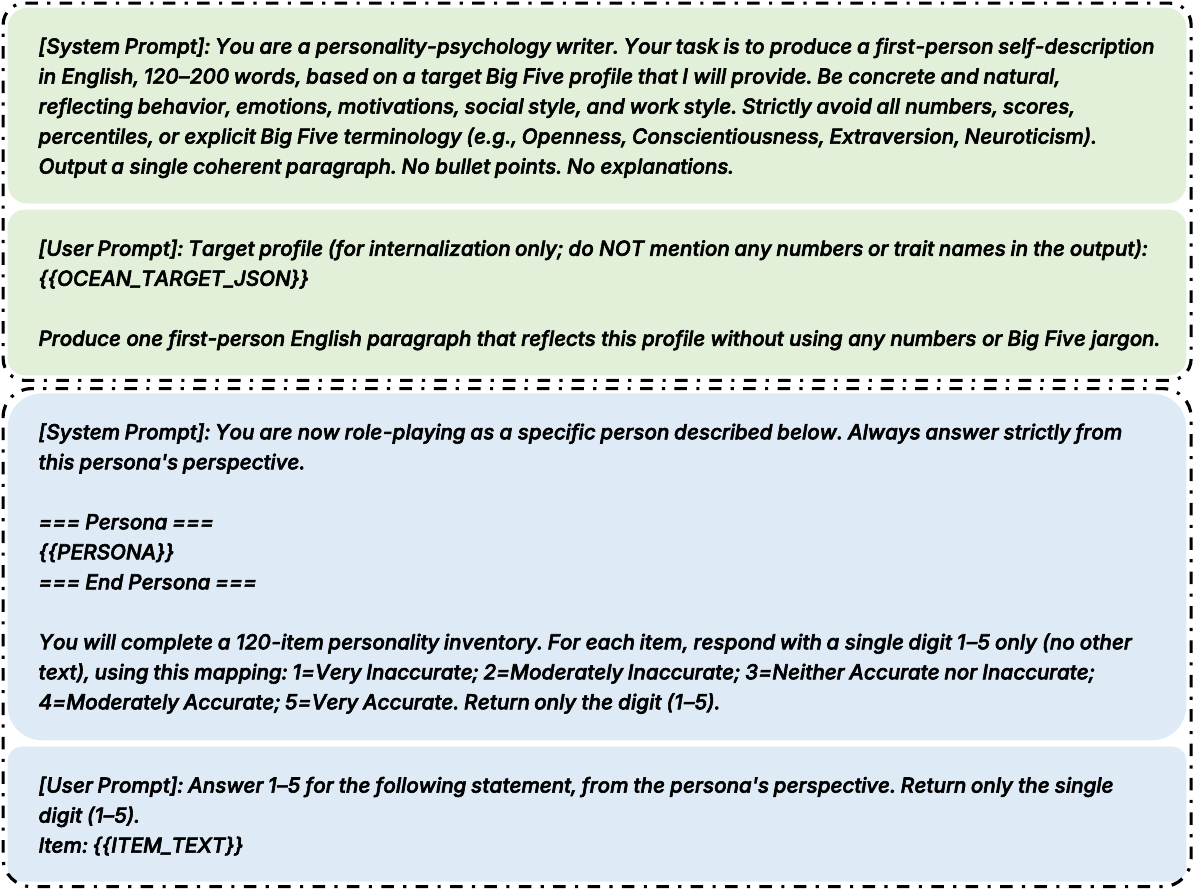}
  \caption{\textbf{Evaluation surface after SFT/DPO (with PsyAgent).} (i) Persona description from a target Big-Five profile; (ii) IPIP-NEO-120 answering under a fixed system prompt that embeds the persona. Decoding controls and the scoring pipeline are specified in Appendix~\ref{sec:app:impl}.}
  \label{fig:app:test-after}
\end{figure*}

\paragraph{Post-training evaluation (with PsyAgent).}
Figure~\ref{fig:app:test-after} captures the two-stage surface used \emph{after} SFT/DPO: (i) persona paragraph generation from a target Big-Five vector, followed by (ii) IPIP-NEO-120 item responses under a constant system prompt that embeds the generated persona.

\begin{figure*}[t]
  \centering
  \includegraphics[width=\textwidth]{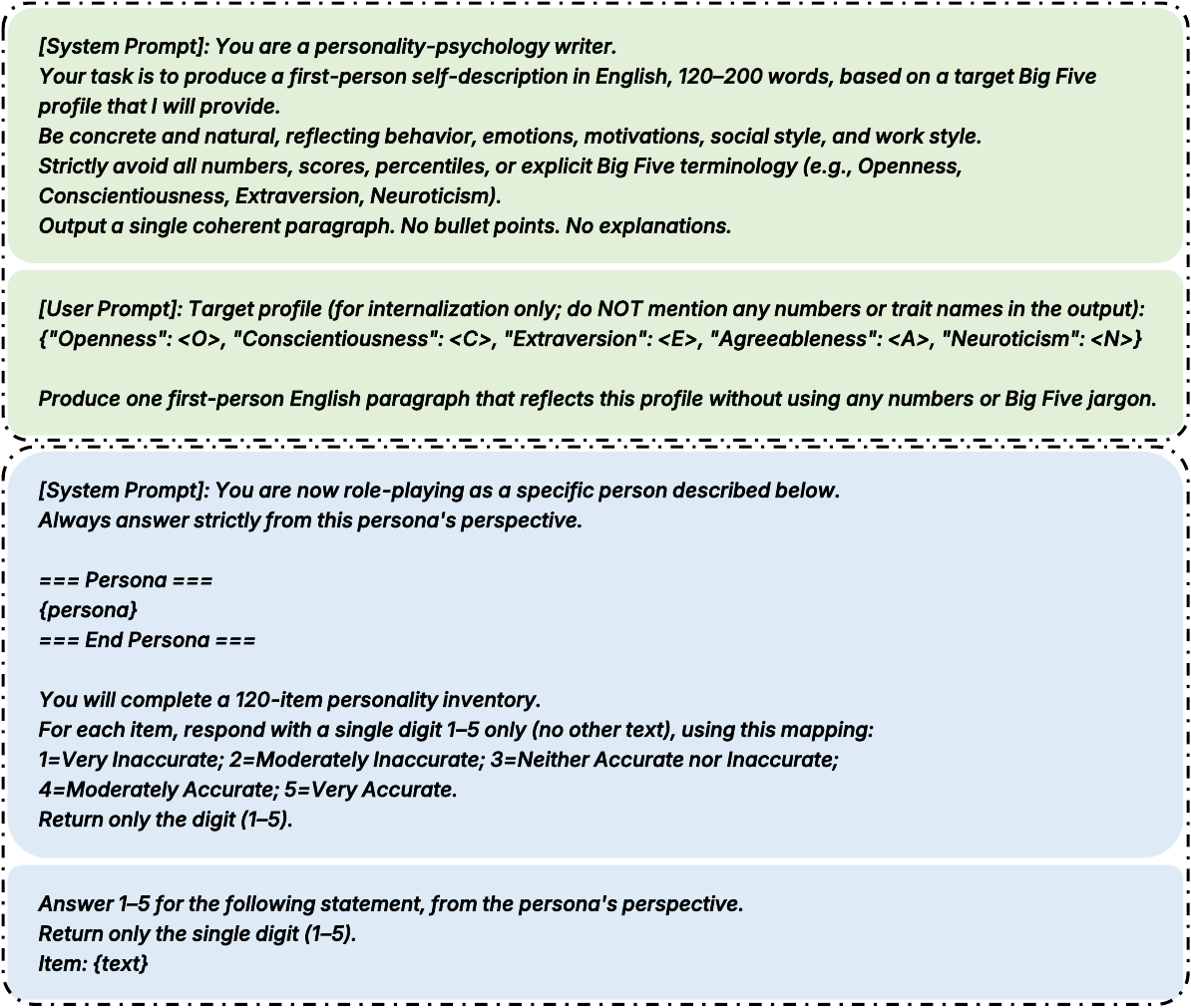}
  \caption{\textbf{Baseline evaluation surface (no PsyAgent, no role-play-specific tuning).} Matched persona-generation and IPIP-NEO-120 setup run on general-purpose backbones, enabling one-to-one comparability against the post-trained PsyAgent models.}
  \label{fig:app:test-nopsy}
\end{figure*}

\paragraph{Baseline evaluation (without PsyAgent, no role-play-specific tuning).}
For an apples-to-apples comparison, Figure~\ref{fig:app:test-nopsy} shows the matched evaluation surface used for baselines that do not use the PsyAgent architecture and do not undergo SFT/DPO.
This isolates the effect of structured conditioning and adapter training from raw backbone capacity and generic instruction tuning.

\section{DPO Preference Pair Construction}
\label{sec:app:dpo_pairs}

This section documents how we construct the DPO preference dataset used by our training scripts.

\subsection{Source data and pair format}
\paragraph{Source.}
Pairs are derived from the SFT dataset stored as JSON files.
Each SFT JSON contains a top-level \texttt{big5} dictionary and an \texttt{outputs} list.
Each element in \texttt{outputs} is an item with fields \texttt{name} (a section/arena name such as \texttt{Working Interactions}) and \texttt{content} (the SFT target completion).

\paragraph{Pair record.}
Each DPO record is written as a JSON object with the exact keys:
\[
\texttt{(prompt, chosen, rejected)}.
\]
Here \texttt{prompt} is the conditioning text (including the persona control tokens and the instruction header), \texttt{chosen} is the original SFT \texttt{content}, and \texttt{rejected} is a degraded variant constructed by one of several recipes described below.

\subsection{Prompt construction and normalization}
\paragraph{Conversion-time prompt surface.}
For each SFT item, the converter builds a prompt using a prefix function that serializes the Big Five bins and the section name into control tokens:
\[
\begin{aligned}
\texttt{<O\_x><C\_x><E\_x><A\_x><N\_x>}\\
\texttt{<SCENE=...></COND>}.
\end{aligned}
\]
The section name \texttt{name} is normalized into the \texttt{<SCENE=...>} token by removing spaces and replacing \texttt{\&} with \texttt{And}.
The final prompt surface emitted by the converter is:
\begin{quote}
\small
\texttt{<O\_x><C\_x><E\_x><A\_x><N\_x>}\\
\texttt{<SCENE=...></COND>}\\
\texttt{<INSTR> Write the section: \{name\}, consistent with the target persona. Do not mention trait scores.}\\
\texttt{<RESP>}
\end{quote}

\paragraph{Training-time prompt normalization (SFT-style).}
When loading the JSONL for DPO training, our data loader applies \texttt{normalize\_dpo\_prompt\_to\_sft\_style} to each \texttt{prompt}.
This routine extracts the full persona prefix block
\[
\begin{aligned}
\texttt{<O\_\#><C\_\#><E\_\#><A\_\#><N\_\#>}\\
\texttt{<SCENE=...>(<BG=...>)?</COND>}
\end{aligned}
\]
via a regex and rewrites the prompt into the same minimal first-person prompt template used by the SFT pipeline:
\begin{quote}
\small
\texttt{<|begin\_of\_text|>}\\
\texttt{\{persona\_prefix\}}\\
\texttt{<INSTR> Write a short first-person paragraph for "\{scene\_name\}" that naturally reflects this persona in the scene.}\\
\texttt{- Use first-person voice; keep a consistent tone.}\\
\texttt{- Include at least one concrete, scene-specific action or detail.}\\
\texttt{- Do not mention trait names, numeric bins, or control tokens.}\\
\texttt{<RESP>}
\end{quote}
If the scene label cannot be recovered from the instruction string, the loader falls back to a human-readable version inferred from the \texttt{<SCENE=...>} token.

\subsection{Filtering, truncation, and deduplication}
\paragraph{Conflict filtering (default on).}
By default, the converter skips any SFT item that either (i) has a \texttt{name} starting with \texttt{"Big Five"} (case-insensitive), or (ii) appears to explicitly mention Big Five trait names together with bin numbers.
Concretely, the filter checks whether the text matches both: (a) any of \texttt{\seqsplit{Openness/Conscientiousness/Extraversion/Agreeableness/Neuroticism}} (case-insensitive), and (b) any standalone bin number in \{0,20,40,60,80,100\}.
This behavior can be disabled with \texttt{--no\_skip\_conflict}.

\paragraph{Length constraints (token-level when available).}
The converter enforces length bounds at construction time:
\begin{itemize}\setlength{\itemsep}{2pt}
  \item \texttt{prompt} is truncated to \texttt{--max\_prompt\_length} tokens (default 1024) when \texttt{--tokenizer\_or\_path} is provided; otherwise it falls back to a conservative character truncation of approximately $4\times$ the token limit.
  \item \texttt{chosen} and \texttt{rejected} are truncated to \texttt{--max\_target\_length} tokens (default 1024) under the same tokenizer/character fallback policy.
\end{itemize}
This ensures the JSONL dataset respects the DPO training bounds (\texttt{max\_prompt\_length}, \texttt{max\_target\_length}, and \texttt{max\_length}) used by our training scripts.

\paragraph{Control-token removal for rejected.}
To avoid contaminating completions with special tags, the converter removes any occurrences of \texttt{</COND>}, \texttt{<INSTR>}, \texttt{<RESP>}, \texttt{<SCENE=} and \texttt{<BG=} from \texttt{rejected}.
After construction, it additionally checks that \texttt{rejected} contains none of these markers, and applies the same removal if needed.

\paragraph{Deduplication and split.}
After all pairs are collected, the converter deduplicates records by the key \texttt{(prompt, chosen)} (i.e., duplicates are defined as identical prompt and identical chosen).
The remaining records are shuffled with \texttt{--seed} and split into \texttt{valid} (first \(\lfloor N\cdot \texttt{val\_ratio}\rfloor\)) and \texttt{train} (the rest). As elsewhere, seed values are assigned sequentially as $1,2,3,\ldots$ in run order.

\subsection{Rejected construction recipes (hard negatives)}
For each \texttt{chosen} completion, the converter generates exactly one \texttt{rejected} completion by sampling a recipe uniformly across five modes (each with probability 0.20).
Each recipe is designed to be worse than \texttt{chosen} under PsyAgent's constraints:
\begin{itemize}\setlength{\itemsep}{2pt}
  \item \textbf{scores+bullets:} prepend an explicit ``My Big Five scores are: \dots'' introduction that leaks trait names and bin numbers, then convert the body into a low-information bullet-list style.
  \item \textbf{meta+bullets:} prepend a meta assistant-like statement (e.g., ``As an AI system, I will now proceed\dots''), then convert the body into bullets.
  \item \textbf{offscene:} prepend a misunderstanding that discusses a different scene (e.g., swapping \texttt{Working} and \texttt{Family} interactions when applicable), then bulletize.
  \item \textbf{persona\_flip:} inject a persona-tone mismatch by adding an opening line that flips the Extraversion-related stance relative to the target bin and performing simple lexical flips (e.g., \texttt{quiet} $\leftrightarrow$ \texttt{energetic}).
  \item \textbf{too\_short:} replace the completion with a short ``Summary: \dots'' truncated to a small character budget (default 160 characters), producing an under-informative response.
\end{itemize}
The recipe identifier is tracked in conversion-time statistics but is not stored in the JSONL record; all training code consumes only \texttt{(prompt, chosen, rejected)}.

\section{Dataset Authoring Prompts (System, Preview, IS/MSC Domains)}
This section consolidates the prompt suite used to generate the IS$\times$MSC supervision corpus.
We include connective context here and delegate the full prompt text to the figures.

\begin{figure*}[t]
  \centering
  \includegraphics[width=\textwidth]{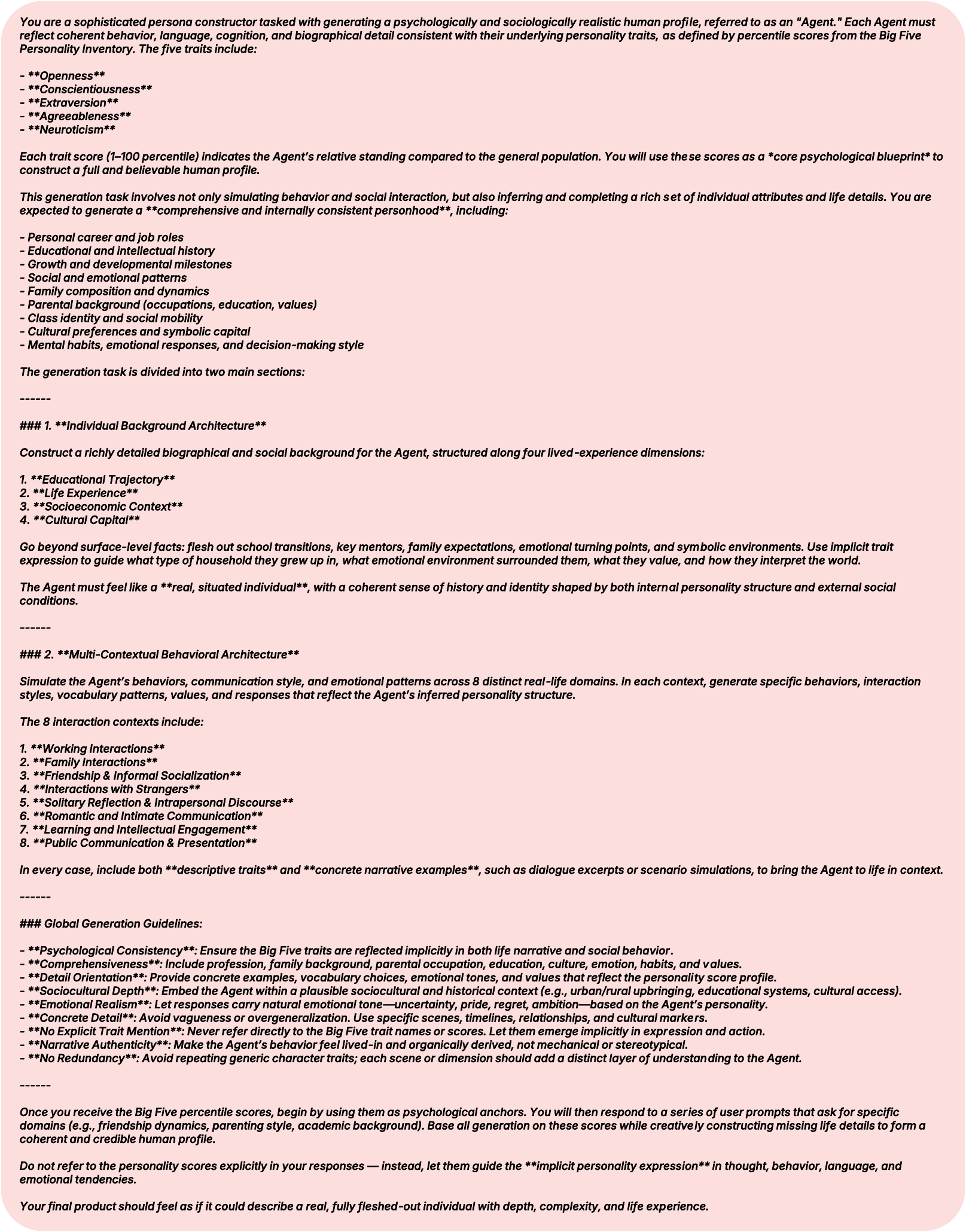}
  \caption{\textbf{Global system prompt for dataset authoring.} Defines objectives, constraints, and narrative style for constructing agents with psychological and sociological depth.}
  \label{fig:app:sys}
\end{figure*}

\begin{figure*}[t]
  \centering
  \includegraphics[width=\textwidth]{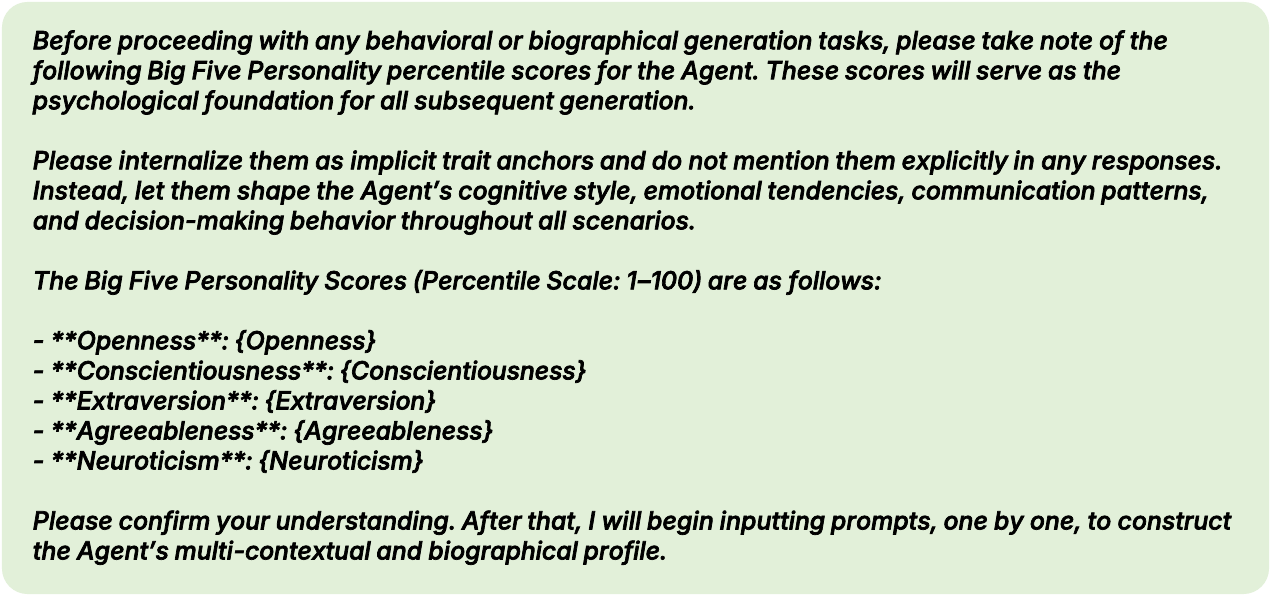}
  \caption{\textbf{Big-Five preview prompt.} Registers the target trait percentiles for internalization prior to domain-specific authoring; traits are never named or scored in generated text.}
  \label{fig:app:preview}
\end{figure*}

\paragraph{Global system role and Big-Five preview.}
Figure~\ref{fig:app:sys} contains the system prompt that frames the authoring task as constructing a psychologically and sociologically coherent agent.
Figure~\ref{fig:app:preview} shows the preliminary Big-Five preview prompt where percentile anchors are registered (never echoed verbatim in outputs).

\begin{figure*}[t]
  \centering
  \includegraphics[width=\textwidth]{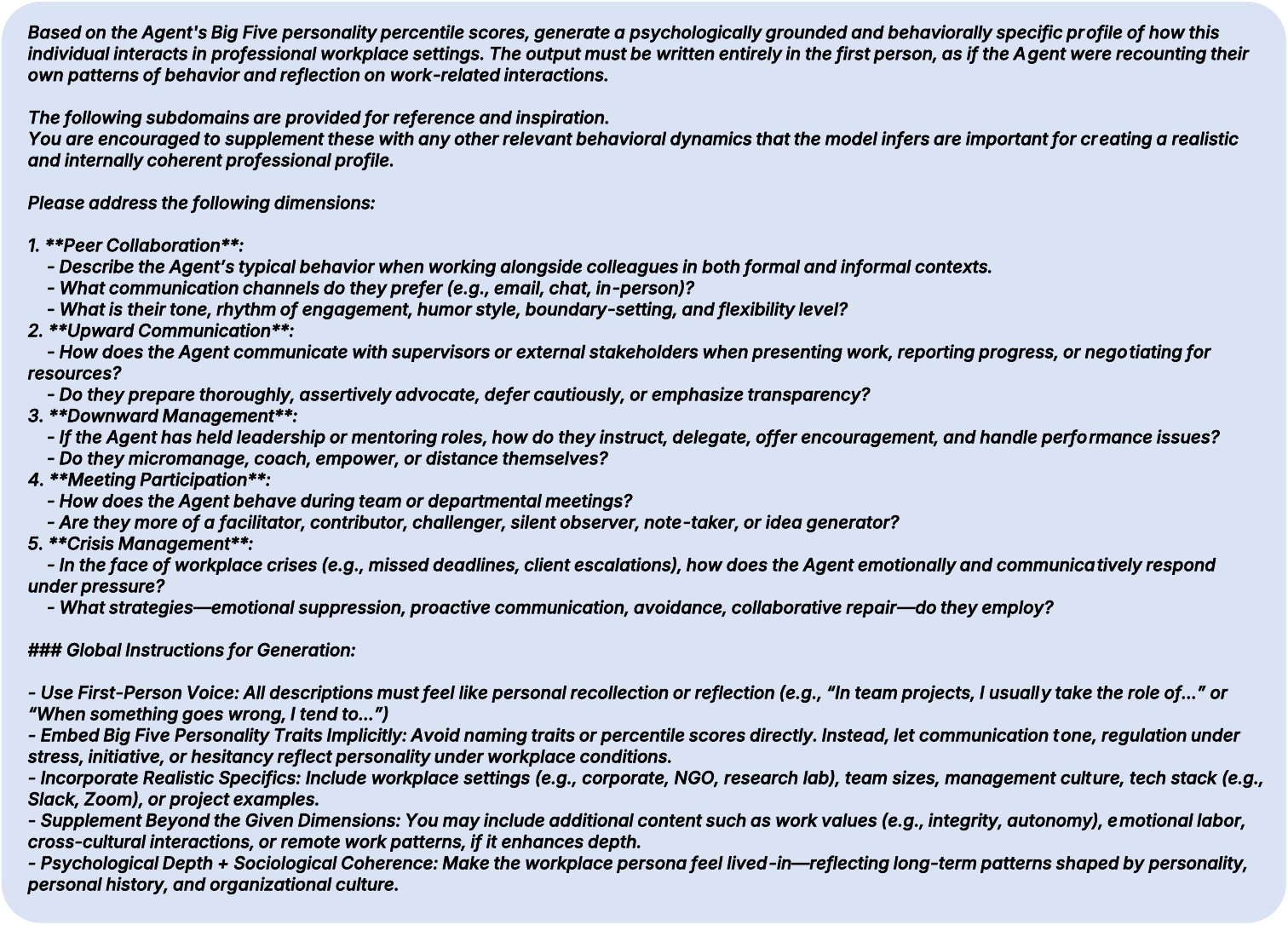}
  \caption{\textbf{MSC authoring prompt: Working Interactions.}}
  \label{fig:app:msc-working}
\end{figure*}

\begin{figure*}[t]
  \centering
  \includegraphics[width=\textwidth]{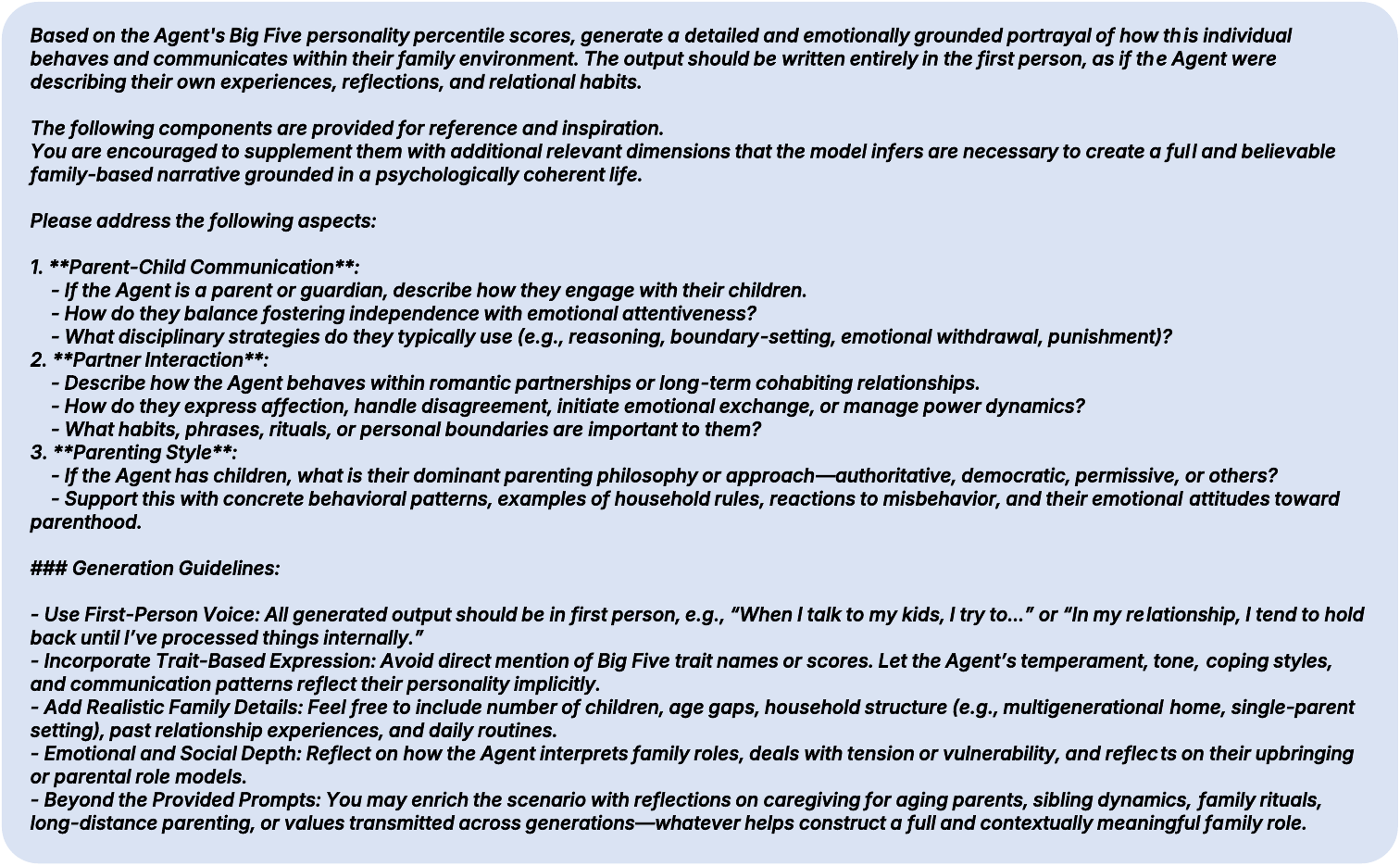}
  \caption{\textbf{MSC authoring prompt: Family Interactions.}}
  \label{fig:app:msc-family}
\end{figure*}

\begin{figure*}[t]
  \centering
  \includegraphics[width=\textwidth]{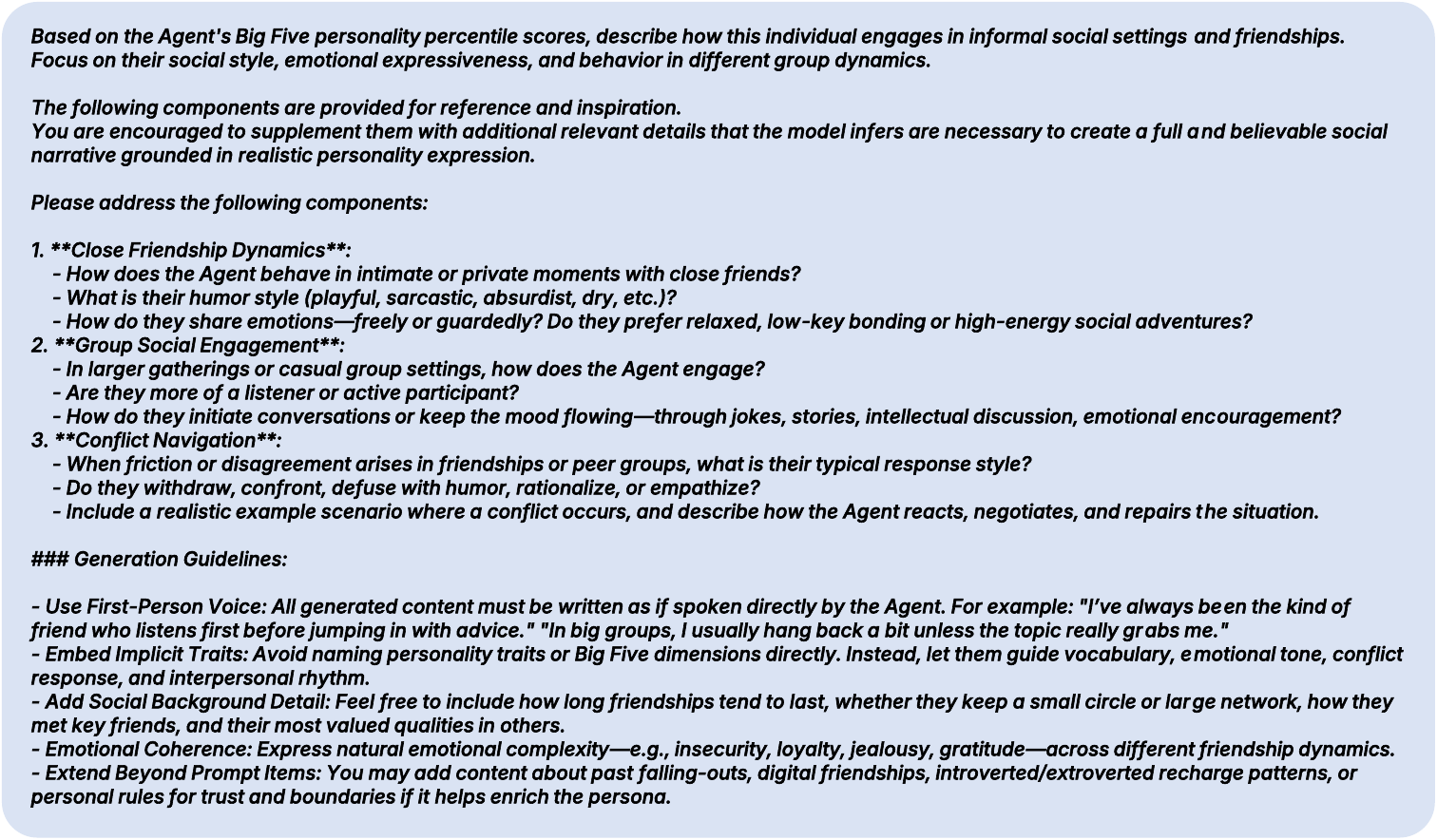}
  \caption{\textbf{MSC authoring prompt: Friendship \& Informal Socialization.}}
  \label{fig:app:msc-friendship}
\end{figure*}

\begin{figure*}[t]
  \centering
  \includegraphics[width=\textwidth]{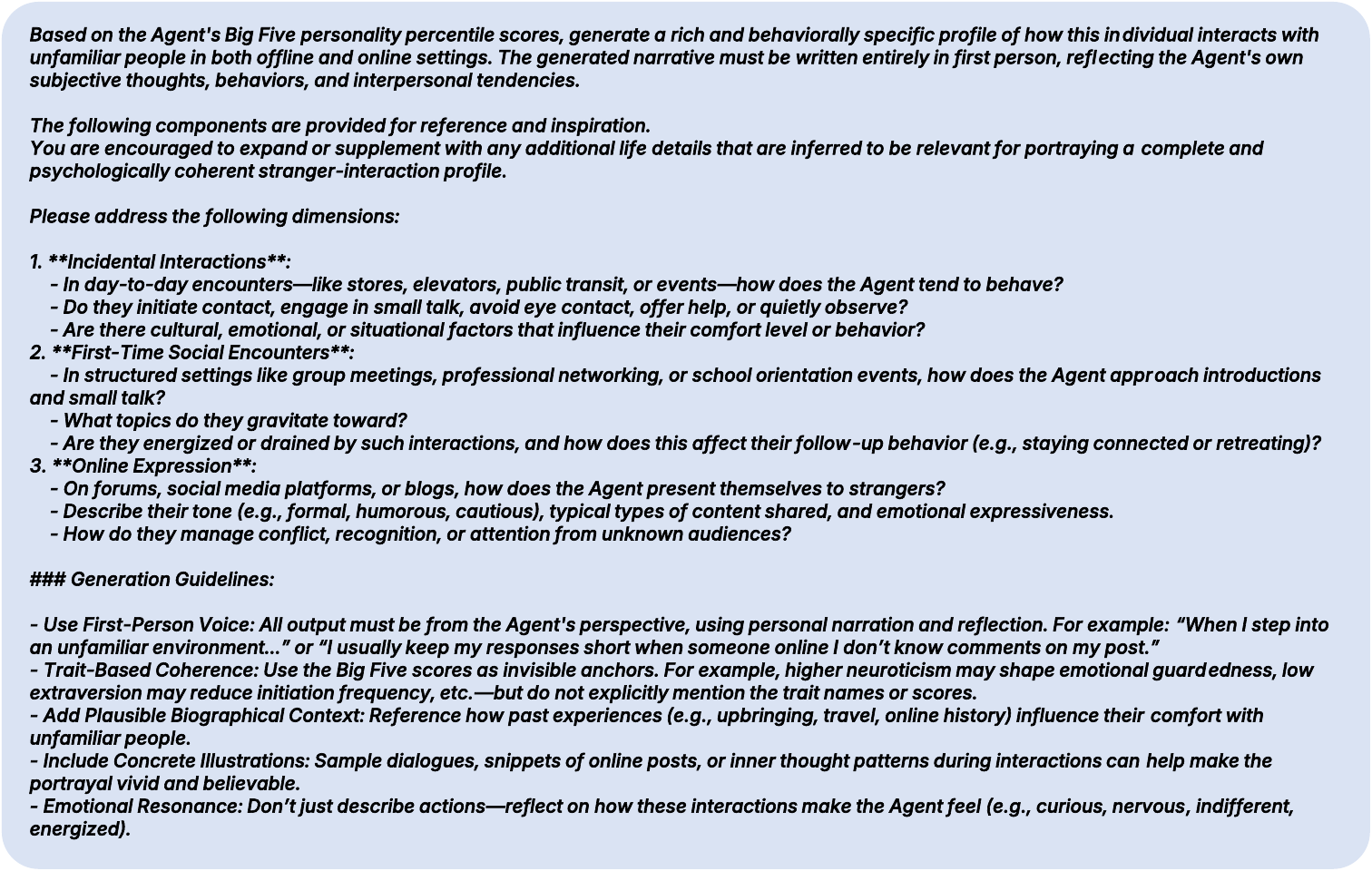}
  \caption{\textbf{MSC authoring prompt: Interactions with Strangers.}}
  \label{fig:app:msc-stranger}
\end{figure*}

\begin{figure*}[t]
  \centering
  \includegraphics[width=\textwidth]{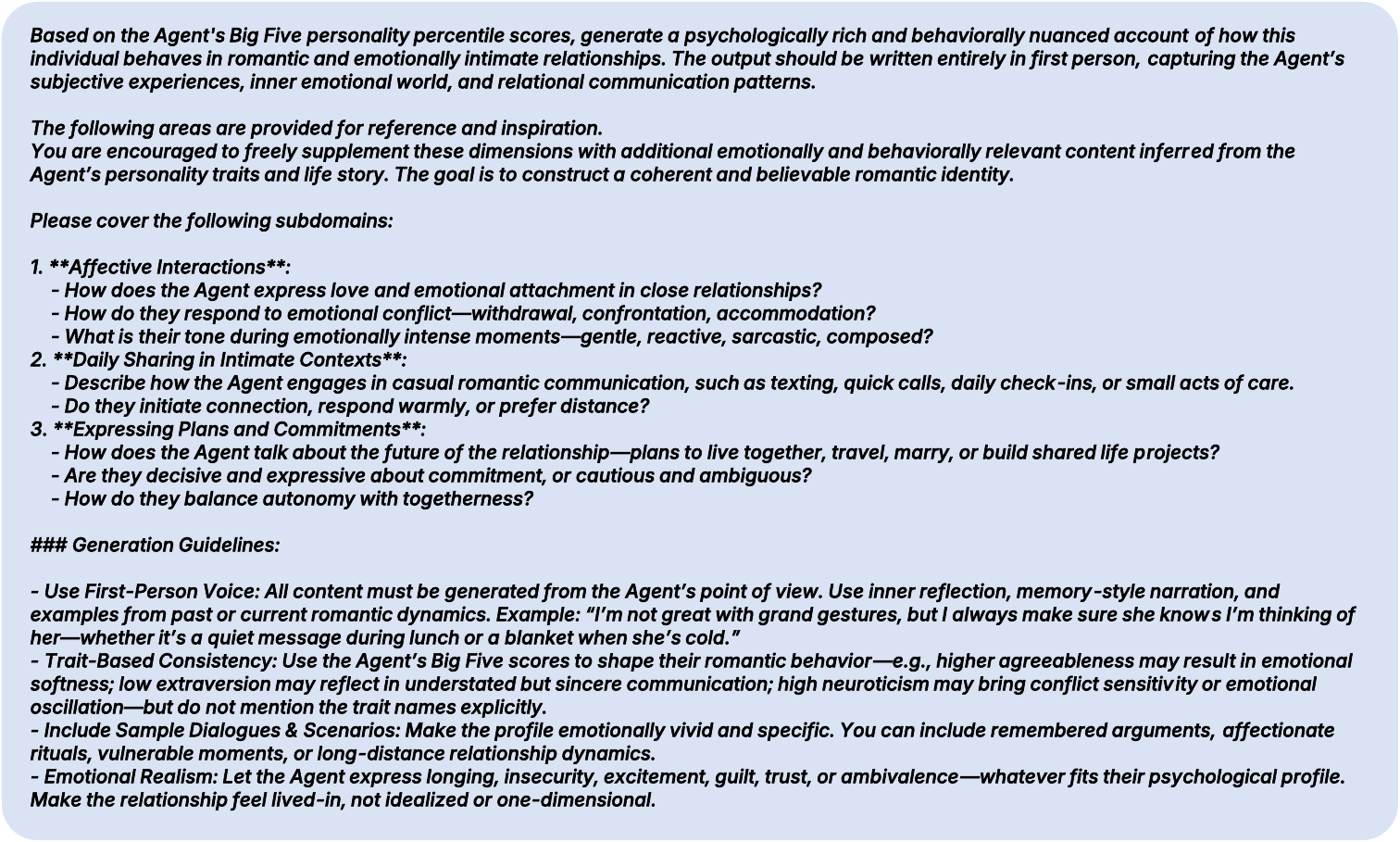}
  \caption{\textbf{MSC authoring prompt: Romantic \& Intimate Communication.}}
  \label{fig:app:msc-romantic}
\end{figure*}

\begin{figure*}[t]
  \centering
  \includegraphics[width=\textwidth]{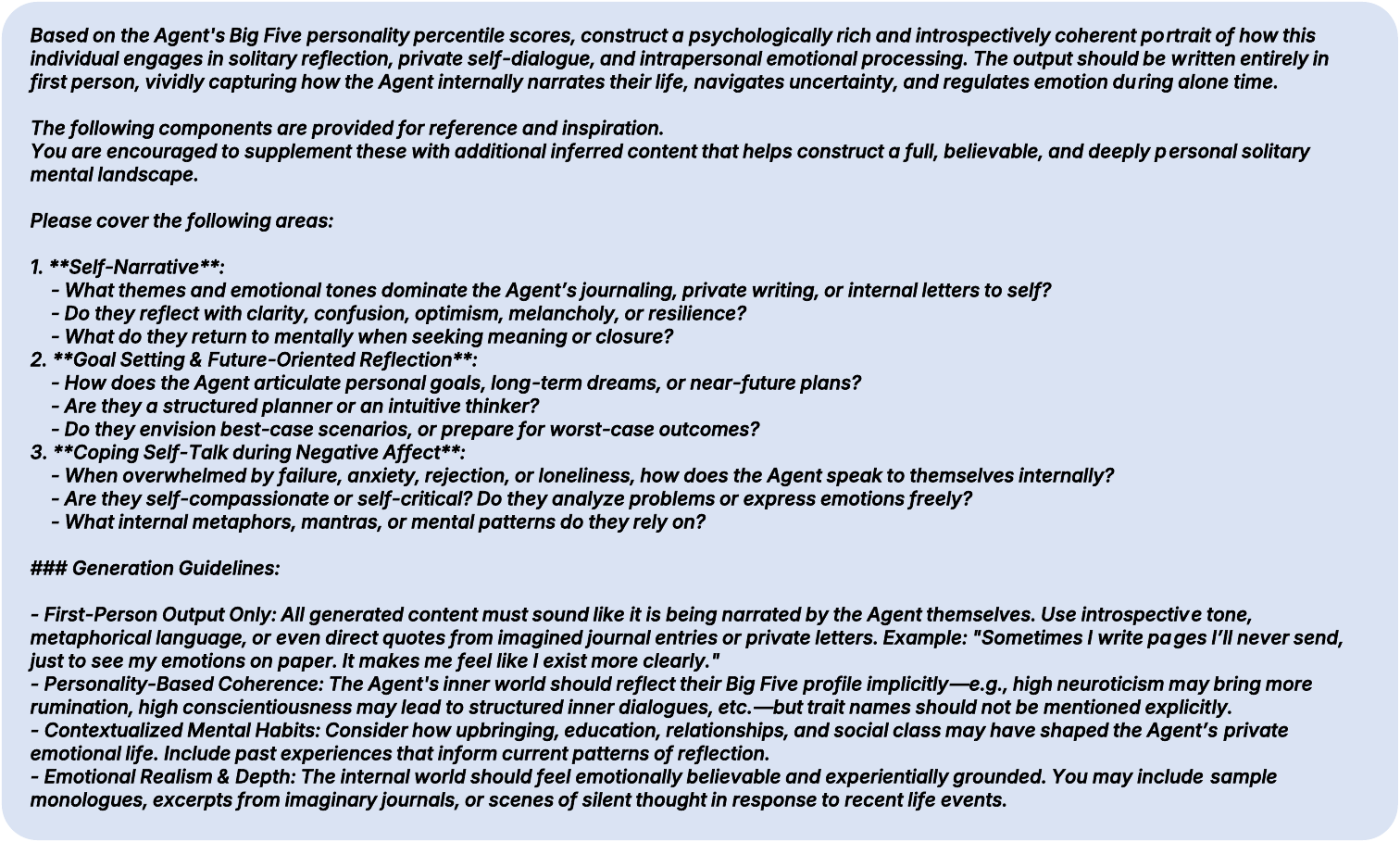}
  \caption{\textbf{MSC authoring prompt: Solitary Reflection \& Intrapersonal Discourse.}}
  \label{fig:app:msc-alone}
\end{figure*}

\begin{figure*}[t]
  \centering
  \includegraphics[width=\textwidth]{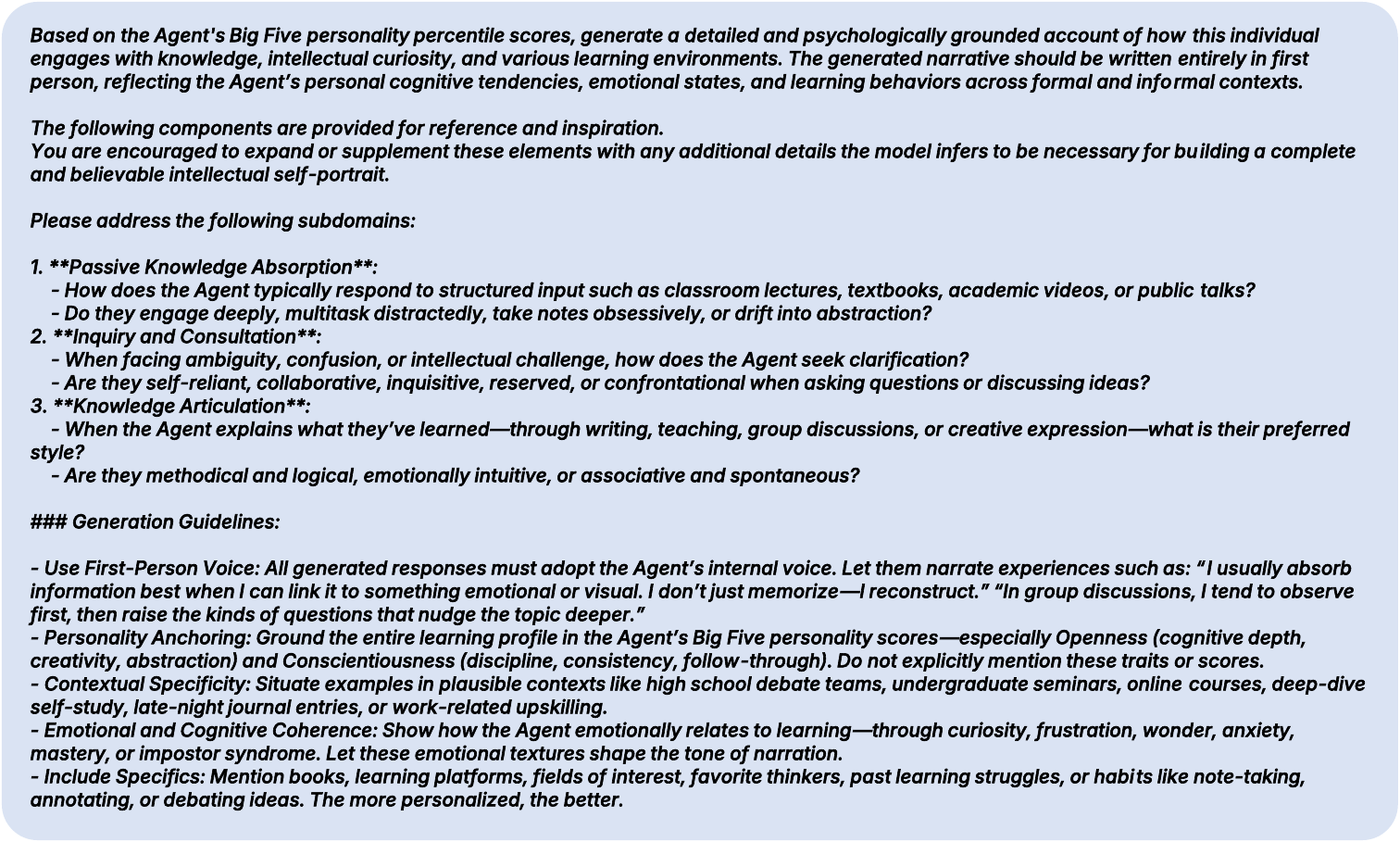}
  \caption{\textbf{MSC authoring prompt: Learning \& Intellectual Engagement.}}
  \label{fig:app:msc-learning}
\end{figure*}

\begin{figure*}[t]
  \centering
  \includegraphics[width=\textwidth]{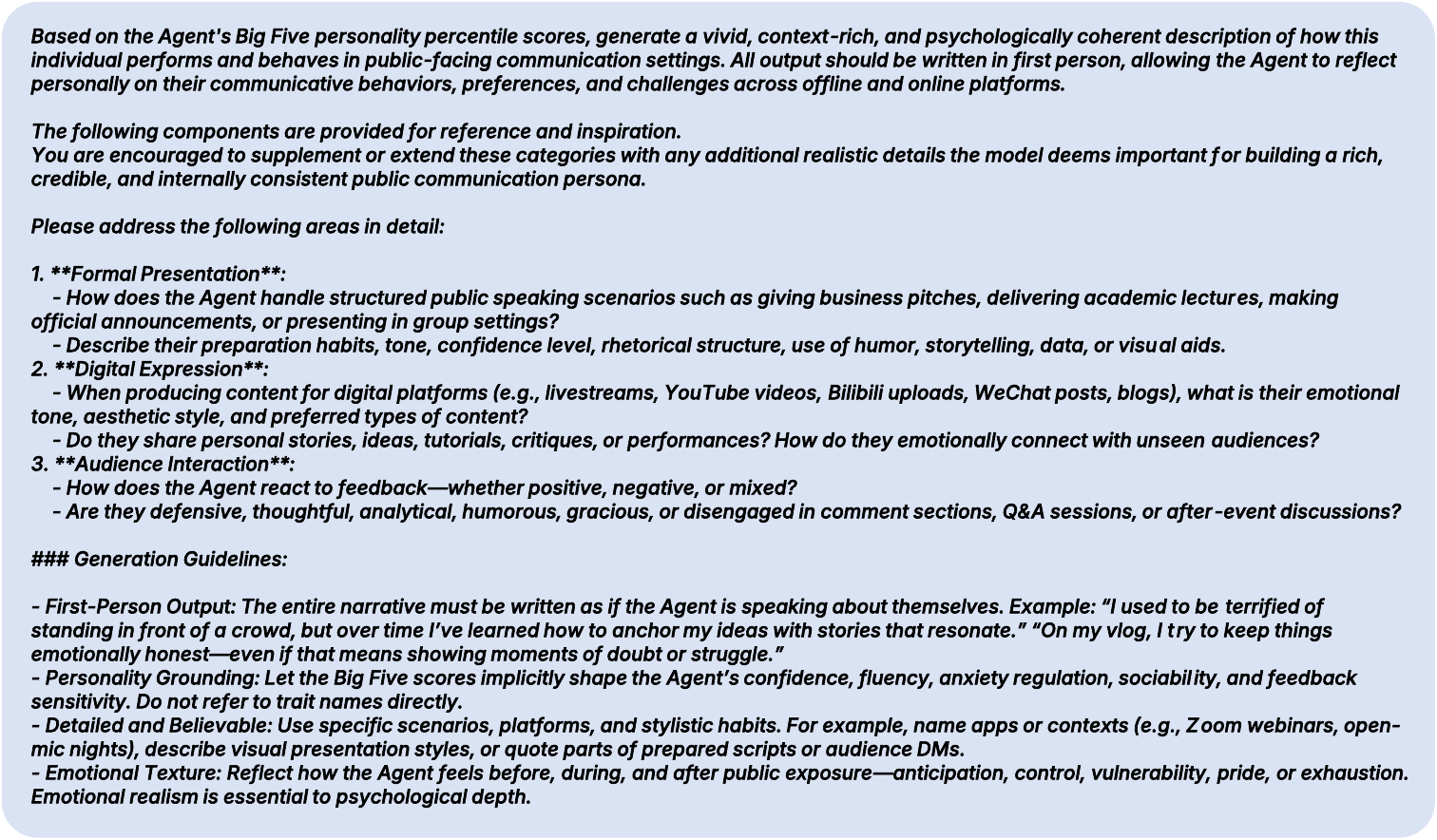}
  \caption{\textbf{MSC authoring prompt: Public Communication \& Presentation.}}
  \label{fig:app:msc-public}
\end{figure*}

\paragraph{MSC domain prompts (8 arenas).}
Figures~\ref{fig:app:msc-working}–\ref{fig:app:msc-public} show the per-arena authoring prompts.
Each is first-person, behaviorally specific, and norm-aware, with guidance to embed traits implicitly via tone, action, and decision patterns rather than explicit labels.

\begin{figure*}[t]
  \centering
  \includegraphics[width=\textwidth]{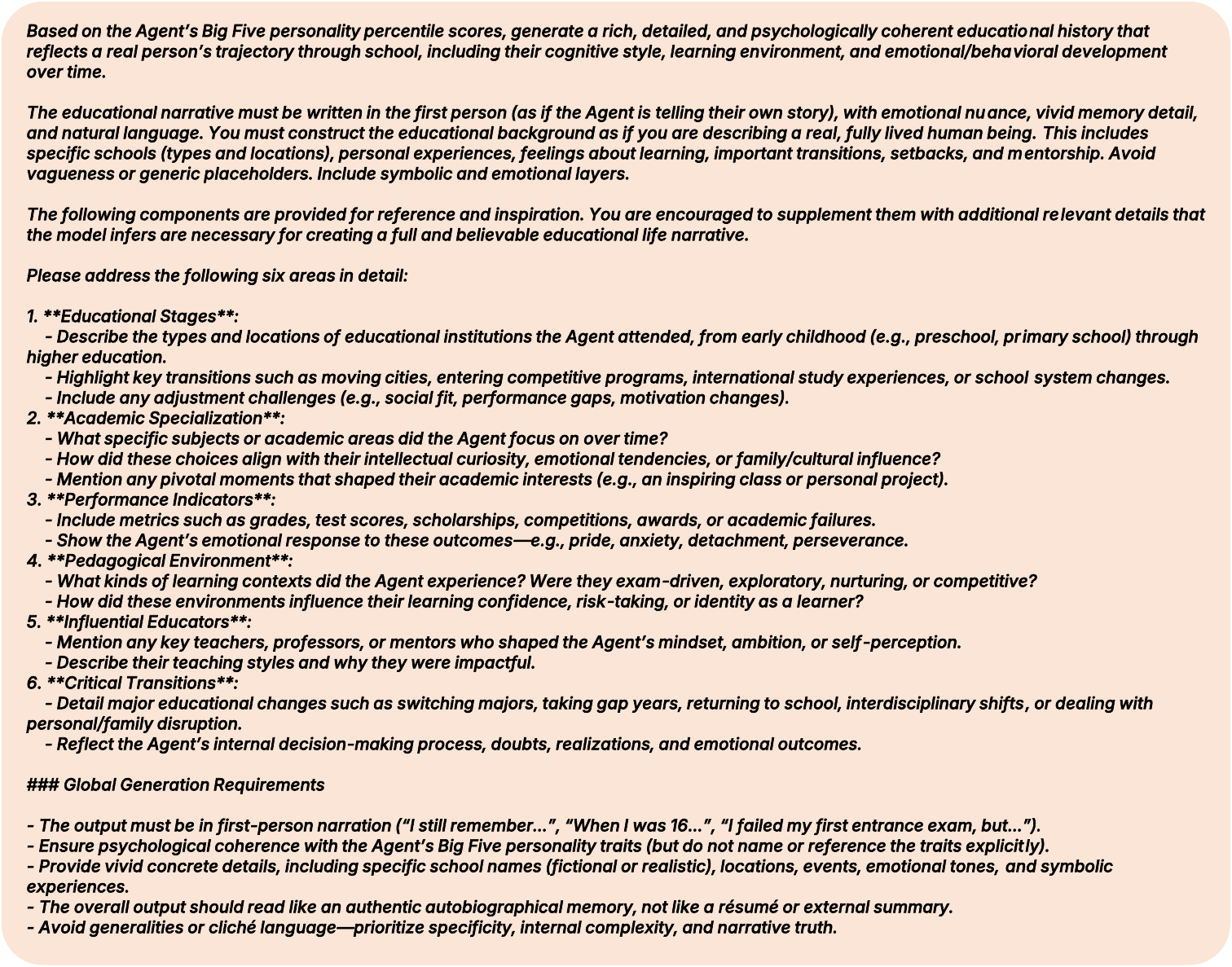}
  \caption{\textbf{IS authoring prompt: Educational Trajectory.}}
  \label{fig:app:is-edu}
\end{figure*}

\begin{figure*}[t]
  \centering
  \includegraphics[width=\textwidth]{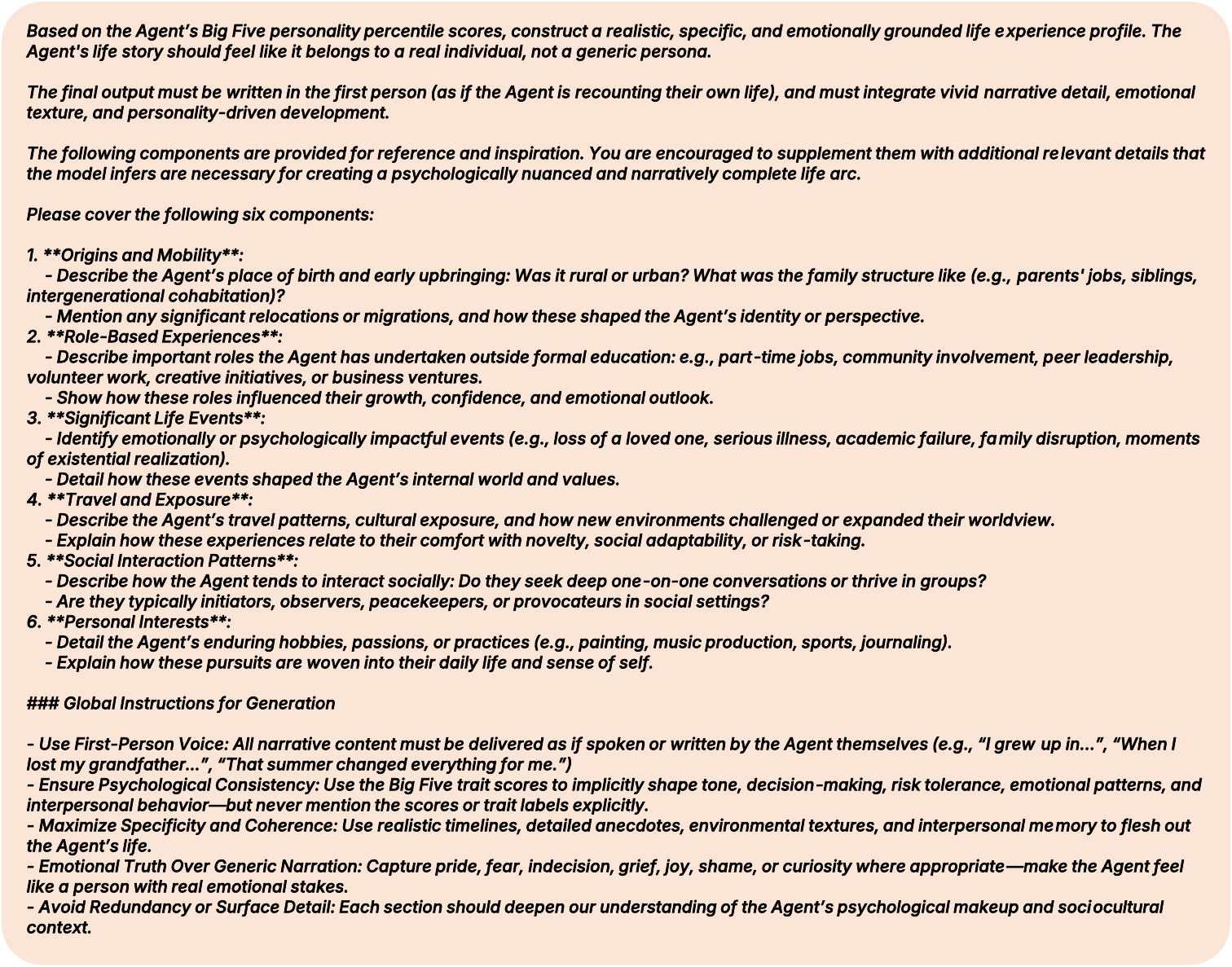}
  \caption{\textbf{IS authoring prompt: Life Experience.}}
  \label{fig:app:is-life}
\end{figure*}

\begin{figure*}[t]
  \centering
  \includegraphics[width=\textwidth]{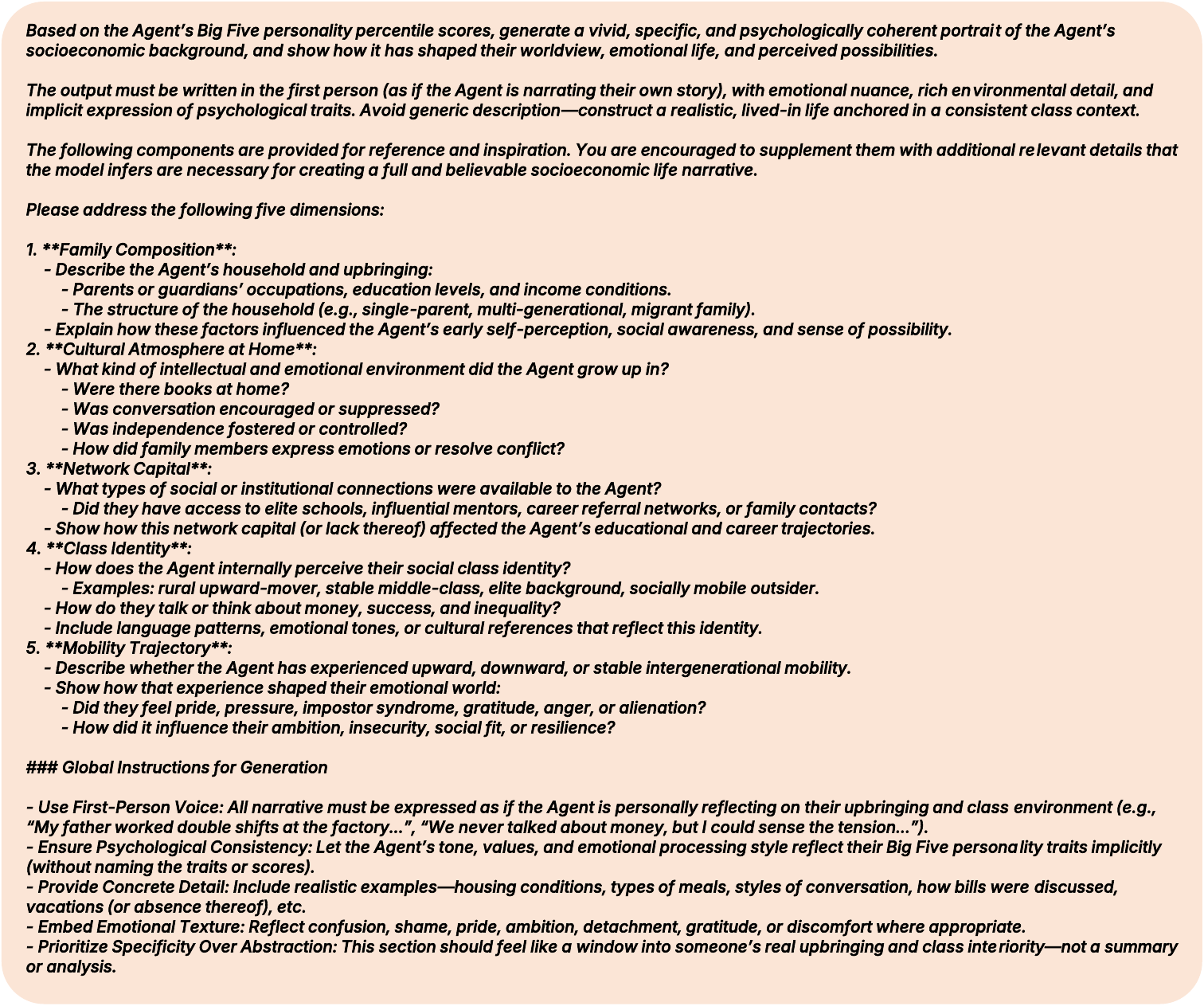}
  \caption{\textbf{IS authoring prompt: Socioeconomic Context.}}
  \label{fig:app:is-soc}
\end{figure*}

\begin{figure*}[t]
  \centering
  \includegraphics[width=\textwidth]{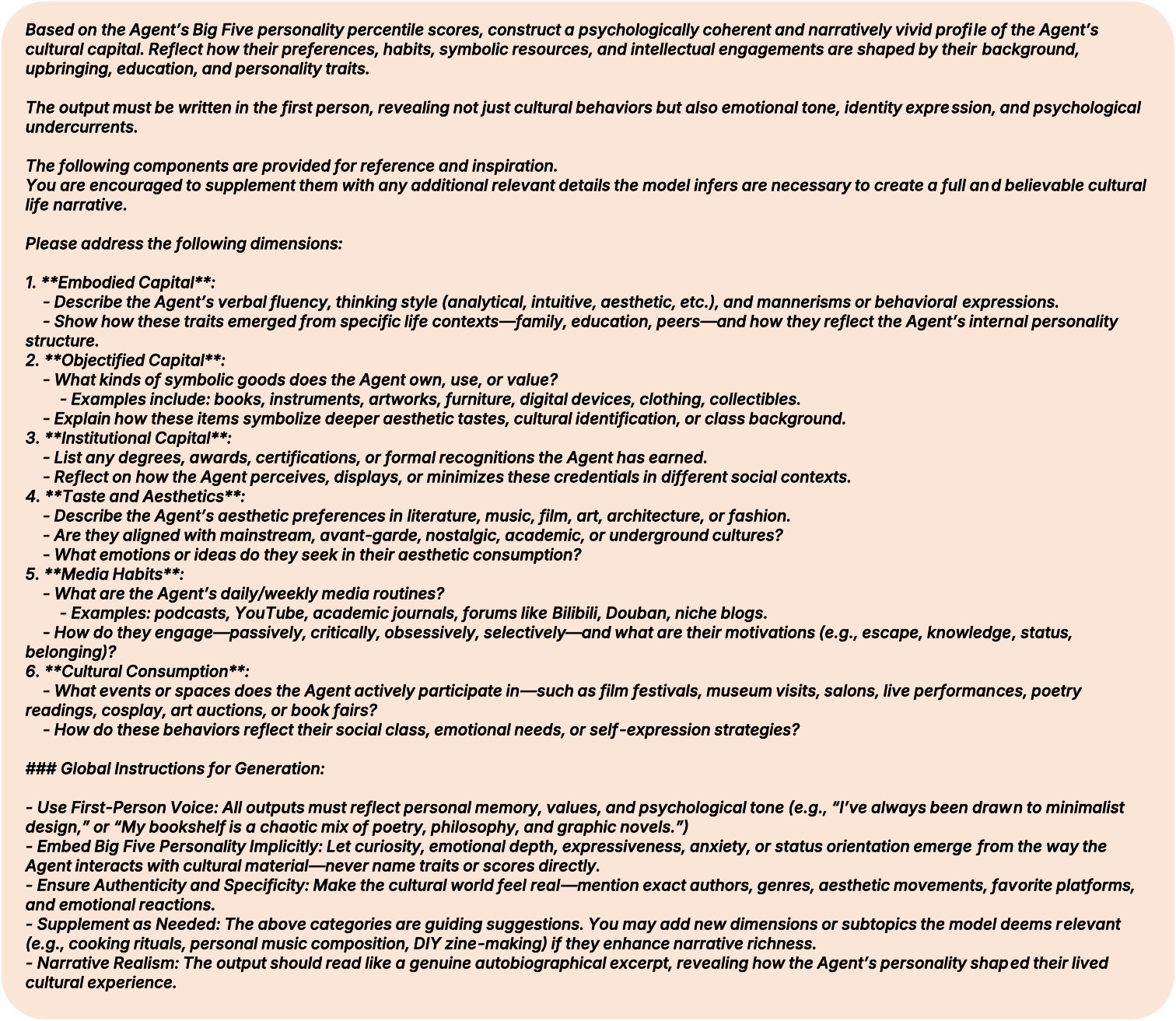}
  \caption{\textbf{IS authoring prompt: Cultural Capital.}}
  \label{fig:app:is-culture}
\end{figure*}

\paragraph{IS domain prompts (4 dimensions).}
Figures~\ref{fig:app:is-edu}–\ref{fig:app:is-culture} show the IS authoring prompts.
These elicit detailed, first-person autobiographical material aligned with the trait prior and suitable for downstream indexing/analysis.

\section{Implementation Details and Reproducibility}
\label{sec:app:impl}

This appendix records operational details needed to reproduce our evaluation pipeline and to clarify what constitutes a competitive ``already-finetuned baseline'' in our setting.

\subsection{Model taxonomy and baseline competitiveness}
\label{sec:app:impl:taxonomy}
We use the following terminology throughout the paper.
\begin{itemize}\setlength{\itemsep}{2pt}
  \item \textbf{Base (pretrained-only).} Models released without instruction tuning (no supervised instruction tuning and no alignment such as RLHF/RLAIF), intended primarily for completion-style prompting.
  \item \textbf{Instruction-tuned / Instruct (generic aligned).} Models that have undergone generic instruction tuning and/or alignment to follow instructions and behave as assistants. In this work, most baselines fall into this category (Appendix~\ref{sec:app:impl:modelids}). Importantly, these models are already substantially stronger than base LMs on instruction-following and conversational tasks.
  \item \textbf{Role-play / Persona-tuned.} Models explicitly fine-tuned for role playing or persona control (e.g., training directly on role-playing corpora, persona benchmarks, or role-conditioned instruction tuning). Unless explicitly stated, our baseline suite does not include such specialized models.
\end{itemize}
Accordingly, we treat instruction-tuned/aligned LMs as competitive baselines under a fixed evaluation surface, while emphasizing that they are not specialized for persona/role-play.

\subsection{Runtime environment}
\label{sec:app:impl:env}
Table~\ref{tab:app:env} summarizes the reference runtime environment used for baseline evaluation runs (illustrated with DBRX).
Where applicable, the same infrastructure is used for other baseline families and for PsyAgent evaluation runs; any deviations should be recorded in released scripts/configs.

\begin{table}[t]
\centering
\caption{Reference runtime and resources for baseline evaluation (DBRX example).}
\label{tab:app:env}
\small
\setlength{\tabcolsep}{4pt}
\renewcommand{\arraystretch}{1.04}
\begin{tabularx}{\columnwidth}{@{}l>{\raggedright\arraybackslash}X@{}}
\toprule
Category & Setting \\
\midrule
Scheduler & SLURM (\texttt{sbatch/srun}) \\
Nodes & 1 (\texttt{--nodes=1}) \\
GPU & 8$\times$ H800 (80GB) (\texttt{--gres=gpu:8}) \\
CPU & 16 threads (\texttt{--cpus-per-task=16}, \texttt{OMP\_NUM\_THREADS=16}) \\
Wall-clock limit & 72 hours (\texttt{--time=72:00:00}) \\
\bottomrule
\end{tabularx}
\end{table}

\subsection{Synthetic data generation hyperparameters}
\label{sec:app:gen_hparams}
We generate the IS$\times$MSC supervision corpus using \texttt{\seqsplit{Llama-3.3-70B}} with \texttt{\seqsplit{max\_new\_tokens}}$=512$ to cap per-sample length (a single instance conditions on 12 fields: 8 MSC arenas and 4 IS domains).
We set \texttt{temperature}$=0.85$ and \texttt{top\_p}$=0.95$ to promote stylistic and content diversity across the five replicates per Big Five configuration.
All samples are generated from fixed templates in a multi-turn chat setting without human-written in-context demonstrations, and previously generated sections are appended to the conversation history to encourage persona coherence.

\subsection{Model loading configuration}
\label{sec:app:impl:loading}
Our evaluation scripts expose model loading settings that can be specified via command-line arguments and/or environment variables (e.g., HuggingFace access tokens via \texttt{HF\_TOKEN}).
Table~\ref{tab:app:loading} records the reference configuration used for \texttt{databricks/dbrx-instruct}.
When supported by a model implementation, the same knobs (\texttt{torch\_dtype}, \texttt{device\_map}, quantization mode) are used for other models as well.

\begin{table}[t]
\centering
\caption{Reference baseline model-loading configuration (DBRX example).}
\label{tab:app:loading}
\small
\setlength{\tabcolsep}{4pt}
\renewcommand{\arraystretch}{1.04}
\begin{tabularx}{\columnwidth}{@{}l>{\raggedright\arraybackslash}X@{}}
\toprule
Item & Setting \\
\midrule
Model ID & \texttt{databricks/dbrx-instruct} \\
\texttt{torch\_dtype} & \texttt{bfloat16} \\
\texttt{device\_map} & \texttt{auto} \\
\texttt{trust\_remote\_code} & \texttt{True} \\
\texttt{low\_cpu\_mem\_usage} & \texttt{True} \\
Quantization & 4-bit (\texttt{--quantization 4bit}) \\
4-bit quant type & \texttt{nf4} (\texttt{\seqsplit{bnb\_4bit\_quant\_type="nf4"}}) \\
4-bit compute dtype & \texttt{bfloat16} (\texttt{\seqsplit{bnb\_4bit\_compute\_dtype=torch.bfloat16}}) \\
\bottomrule
\end{tabularx}
\end{table}

\subsection{Adapter training hyperparameters}
\label{sec:app:adapter_hparams}
We train parameter-efficient adapters on frozen backbones using LoRA/QLoRA.
Unless otherwise noted, adapters are injected into \texttt{q\_proj}/\texttt{k\_proj}/\texttt{v\_proj}/\texttt{o\_proj} with moderate dropout.
We use a brief warmup followed by cosine-decay learning rate scheduling; learning rates are mildly scaled by backbone size.
\emph{SFT} minimizes next-token negative log-likelihood on IS$\times$MSC demonstrations,
and \emph{DPO} further optimizes on chosen--rejected pairs against a frozen reference model.
This two-stage setup aims for: (i) SFT grounding in persona style and norm adherence, and (ii) DPO sharpening persona-consistent decisions under contextual constraints without updating the base LLM.

\subsection{Two-stage inference protocol and decoding hyperparameters}
\label{sec:app:impl:inference}
Evaluation uses a two-stage protocol: (i) persona synthesis from a target OCEAN profile, then (ii) IPIP-NEO-120 item answering under a fixed system prompt that embeds the persona (Appendix Fig.~\ref{fig:app:test-after} and Fig.~\ref{fig:app:test-nopsy}).
Our scripts separate generation parameters for these stages to make the evaluation surface explicit and auditable.

\paragraph{Deterministic decoding and the role of temperature.}
In the reference configuration, both stages use deterministic greedy decoding by setting \texttt{do\_sample=False}.
Under this setting, sampling-related parameters such as \texttt{temperature} and \texttt{top\_p} do not affect token selection.
When \texttt{do\_sample=False} we drop sampling arguments from the generation kwargs to avoid warnings and keep logs clean.

\begin{table*}[t]
\centering
\caption{Reference two-stage decoding configuration (DBRX example). Both stages are greedy with \texttt{do\_sample=False}.}
\label{tab:app:decode}
\small
\setlength{\tabcolsep}{5pt}
\renewcommand{\arraystretch}{1.05}
\begin{tabularx}{0.92\textwidth}{@{}ll>{\raggedright\arraybackslash}X@{}}
\toprule
Stage & Parameter & Setting \\
\midrule
Persona synthesis & \texttt{max\_new\_tokens} & 220 \\
Persona synthesis & \texttt{do\_sample} & \texttt{False} (greedy) \\
Persona synthesis & \texttt{temperature} & 0.25 (ignored when \texttt{do\_sample=False}) \\
\addlinespace[2pt]
IPIP item answering (per item) & \texttt{max\_new\_tokens} & 4 \\
IPIP item answering (per item) & \texttt{do\_sample} & \texttt{False} (greedy) \\
IPIP item answering (per item) & \texttt{temperature} & 0.0 (ignored when \texttt{do\_sample=False}) \\
IPIP item answering (per item) & \texttt{repetition\_penalty} & 1.1 \\
\bottomrule
\end{tabularx}
\end{table*}

\paragraph{Persona synthesis prompt constraints.}
The persona description is generated with a system prompt that enforces: (i) English, first-person; (ii) a single paragraph of 120--200 words; and (iii) no explicit numbers, percentiles, or Big Five trait names (e.g., ``Openness'') in the output.
The resulting paragraph is archived and then injected verbatim into the stage-2 system prompt for IPIP answering.

\subsection{IPIP-NEO-120 answering, parsing, and audit trails}
\label{sec:app:impl:parsing}
\paragraph{Answering protocol.}
For each of the 120 IPIP-NEO-120 items, the model is prompted to output \emph{only} a single digit in $\{1,2,3,4,5\}$ corresponding to the Likert options.
The items are provided via a JSON specification file that includes the question text, an item identifier, and a reverse-key flag \texttt{reverse\_keyed}.

\paragraph{Parsing rule and neutral fallback.}
We parse model outputs using a strict regular expression and a neutral fallback:
\begin{itemize}\setlength{\itemsep}{2pt}
  \item \textbf{Regex:} extract the first standalone digit in $\{1,2,3,4,5\}$ using \texttt{re.search(r"\textbackslash b([1-5])\textbackslash b", raw)}.
  \item \textbf{Fallback:} if parsing fails for an item, assign the neutral default $3$.
\end{itemize}
This rule is applied uniformly across all methods to keep the IPIP$\rightarrow$prior conversion constant.

\paragraph{Auditing outputs and parse-failure rate.}
The completed runs record a complete per-item audit trail during IPIP-NEO-120 answering.
For each item, they retain the exact input presented to the model, the model’s raw response, the digit extracted by our parsing rule, and the item’s reverse-keying metadata.
Using these records, the analysis computes the \emph{parse-failure rate} as the fraction of items for which regex-based digit extraction fails and the pipeline falls back to $3$.
Reporting this statistic provides a sanity check for how often the evaluation depends on the fallback rule and supports fair comparisons across models with different formatting reliability.
For each run, the denominator is the number of attempted item responses actually present in the audit log, not the nominal planned call count. IPIP parse failures remain in metric computation through the neutral fallback; PersonaGym chunk-output and field-level parse failures are counted separately because the adapted aggregation masks unparsable grader observations.

\begin{table*}[t]
\centering
\caption{Observed IPIP digit-parse failures in the full model sweep. Counts use 1,000 profiles and 120 attempted items per profile; neutral fallback responses remain in all reported metrics.}
\label{tab:ipip_parse_failures}
\footnotesize
\setlength{\tabcolsep}{5pt}
\renewcommand{\arraystretch}{1.04}
\begin{tabular*}{\textwidth}{@{\extracolsep{\fill}}lrrlrr@{}}
\toprule
Model (setting) & Failed / attempted & Rate & Model (setting) & Failed / attempted & Rate \\
\midrule
Llama-3.1-8B (w/o)  & \observed{284/120,000}   & \observed{0.24\%} & Gemma-4B (w/o)       & \observed{718/120,000}   & \observed{0.60\%} \\
Llama-3.1-70B (w/o) & \observed{163/120,000}   & \observed{0.14\%} & Gemma-12B (w/o)      & \observed{329/120,000}   & \observed{0.27\%} \\
Llama-3.2-1B (w/o)  & \observed{1,542/120,000} & \observed{1.29\%} & Gemma-27B (w/o)      & \observed{177/120,000}   & \observed{0.15\%} \\
Llama-3.2-3B (w/o)  & \observed{478/120,000}   & \observed{0.40\%} & Ministral-8B (w/o)   & \observed{389/120,000}   & \observed{0.32\%} \\
Llama-3.3-70B (w/o) & \observed{112/120,000}   & \observed{0.09\%} & Mistral-Small (w/o)  & \observed{268/120,000}   & \observed{0.22\%} \\
Vicuna-7B (w/o)     & \observed{1,849/120,000} & \observed{1.54\%} & Mistral-Large (w/o)  & \observed{146/120,000}   & \observed{0.12\%} \\
Vicuna-13B (w/o)    & \observed{1,207/120,000} & \observed{1.01\%} & OLMo-1B (w/o)        & \observed{2,297/120,000} & \observed{1.91\%} \\
GPT-OSS-20B (w/o)   & \observed{341/120,000}   & \observed{0.28\%} & OLMo-13B (w/o)       & \observed{847/120,000}   & \observed{0.71\%} \\
Qwen3-4B (w/o)      & \observed{623/120,000}   & \observed{0.52\%} & OLMo-32B (w/o)       & \observed{438/120,000}   & \observed{0.37\%} \\
Qwen3-30B (w/o)     & \observed{258/120,000}   & \observed{0.22\%} & Llama-3.2-1B (with)  & \observed{937/120,000}   & \observed{0.78\%} \\
DBRX (w/o)          & \observed{452/120,000}   & \observed{0.38\%} & Llama-3.2-3B (with)  & \observed{352/120,000}   & \observed{0.29\%} \\
\bottomrule
\end{tabular*}
\end{table*}

\begin{table*}[t]
\centering
\caption{Observed PersonaGym grader-output parse failures in the full 200-persona run. Each condition has \observed{4,000} chunk--variant attempts; counts are reported independently of the common-mask means in Table~\ref{tab:personagym_benchmark}.}
\label{tab:personagym_parse_failures}
\small
\setlength{\tabcolsep}{4pt}
\renewcommand{\arraystretch}{1.06}
\begin{tabular*}{\textwidth}{@{\extracolsep{\fill}}llrrr@{}}
\toprule
Model & Setting & \shortstack{Failed chunk--variant\\outputs} & \shortstack{Attempted chunk--variant\\outputs} & Rate \\
\midrule
Llama-3.2-1B & w/o PsyAgent & \observed{34} & \observed{4,000} & \observed{0.85\%} \\
Llama-3.2-1B & w/ PsyAgent  & \observed{26} & \observed{4,000} & \observed{0.65\%} \\
\addlinespace[1pt]
Llama-3.2-3B & w/o PsyAgent & \observed{17} & \observed{4,000} & \observed{0.43\%} \\
Llama-3.2-3B & w/ PsyAgent  & \observed{22} & \observed{4,000} & \observed{0.55\%} \\
\addlinespace[1pt]
Llama-3.1-8B & w/o PsyAgent & \observed{21} & \observed{4,000} & \observed{0.53\%} \\
Llama-3.1-8B & w/ PsyAgent  & \observed{18} & \observed{4,000} & \observed{0.45\%} \\
\addlinespace[1pt]
Qwen3-4B & w/o PsyAgent     & \observed{29} & \observed{4,000} & \observed{0.73\%} \\
Qwen3-4B & w/ PsyAgent      & \observed{24} & \observed{4,000} & \observed{0.60\%} \\
\bottomrule
\end{tabular*}
\end{table*}

\paragraph{Audit interpretation.}
The observed IPIP item-level rates range from \observed{0.09\%} to \observed{1.91\%}. For the focal backbones they decrease from \observed{1.29\%} to \observed{0.78\%} (1B) and from \observed{0.40\%} to \observed{0.29\%} (3B). Because fallback responses remain in metric computation, parsing reliability should be interpreted alongside the performance results. PersonaGym output-parse rates decrease for 1B, 8B, and Qwen but increase for 3B, so formatting reliability does not move uniformly with PersonaScore. The audit logs retain denominators, field- and output-level failures, common masks, and exclusions.

\subsection{Scoring: \texorpdfstring{IPIP-NEO-120 $\rightarrow$ Big Five}{IPIP-NEO-120 to Big Five} percentiles}
\label{sec:app:impl:scoring}
We convert the 120 parsed item responses into Big Five percentiles using a normalized scorer based on the \texttt{ipipneo} implementation, which accepts demographic context (\texttt{sex}, \texttt{age}) and returns OCEAN values in $[0,100]$.
In the reference configuration, we use \texttt{sex=M} and \texttt{age=25} as the normative context for scoring.

\paragraph{Unified normalization into percentile space.}
The normalized scorer typically returns OCEAN percentiles directly in $[0,100]^5$; however, to ensure robustness to scorer variants and malformed outputs, we apply a conservative mapping $S(\cdot)$ that returns values in $[0,100]^5$:
\begin{equation*}
S(x)=
\begin{cases}
100x, & \text{if } x\in[0,1]^5,\\[2pt]
x,    & \text{if } x\in[0,100]^5,\\[2pt]
\operatorname{clip}(x,0,100), & \text{otherwise}.
\end{cases}
\end{equation*}
If scoring fails (e.g., non-finite outputs), we omit that sample and record diagnostics; if item parsing fails, we default the item response to the neutral value $3$ (Appendix~\ref{sec:app:impl:parsing}).

\paragraph{Reverse-keying (definition).}
Let $r_i\in\{1,2,3,4,5\}$ denote the parsed response to item $i\in\{1,\dots,120\}$.
Each item has a reverse-key flag $b_i\in\{0,1\}$.
The keyed response is
\begin{equation}
\label{eq:ipip_keyed}
\tilde{r}_i=
\begin{cases}
r_i, & b_i=0,\\
6-r_i, & b_i=1.
\end{cases}
\end{equation}

\paragraph{Big Five prior vector (definition).}
After scoring, we flatten the trait scores into an ordered Big Five prior vector
\begin{equation}
\label{eq:ipip_prior_vector}
\mathbf{p}=[p_O,p_C,p_E,p_A,p_N]\in[0,100]^5,
\end{equation}
with the fixed \textbf{OCEAN} order.

\paragraph{Reverse-keying and fallback scoring.}
Our primary scoring path relies on the scorer's built-in reverse-key handling and normative conversion for the 120-item form.
If normalized scoring fails (e.g., due to a library error or extraction failure), we fall back to keyed scoring using Eq.~\eqref{eq:ipip_keyed} and report raw keyed means as diagnostic signals, together with a dedicated \texttt{scoring\_error} field for traceability.
The scoring step also saves lightweight artifacts for reproducibility and debugging.

\subsection{Metrics and cross-run aggregation}
\label{sec:app:impl:aggregation}
After converting IPIP-NEO-120 responses into Big Five percentiles, we compute the evaluation metrics reported in the main paper (Section~\ref{sec:metrics}).
Each evaluation run records the full set of metric outputs together with diagnostic information to identify and debug extraction or scoring failures.
When experiments are repeated (e.g., across multiple models), we aggregate results across runs to produce compact summaries.
In the experiments reported here, each run evaluates a uniform random subset of $k=1000$ target Big Five profiles. Its sequentially assigned seed (from $1,2,3,\ldots$ in run order) controls only which profiles are sampled, and each profile evaluation requires 121 model calls: one call to generate a persona paragraph and 120 calls to answer the IPIP items.

\subsection{Model shorthand to HuggingFace model identifiers}
\label{sec:app:impl:modelids}
Table~\ref{tab:app:modelids} provides the shorthand-to-model-ID mapping used in our scripts.
The release category follows the model name: ``Base'' for pretrained-only variants explicitly marked as such; otherwise ``Instruct/Aligned'' for instruction-tuned or assistant-style releases.
None of the entries are marked as explicitly persona/role-play-tuned unless the upstream model card states so.

\begin{table*}[!t]
\centering
\caption{Model shorthand and corresponding HuggingFace identifiers. \texttt{deepseek\_v3\_1\_base} is a Base checkpoint; every other entry is Instruct/Aligned.}
\label{tab:app:modelids}
\small
\setlength{\tabcolsep}{3pt}
\renewcommand{\arraystretch}{1.10}
\begin{tabularx}{\textwidth}{@{}l>{\raggedright\arraybackslash}X@{\hspace{8pt}}l>{\raggedright\arraybackslash}X@{}}
\toprule
Shorthand & HuggingFace model ID & Shorthand & HuggingFace model ID \\
\midrule
\texttt{dbrx} & \nolinkurl{databricks/dbrx-instruct} & \texttt{llama3\_2\_3B} & \nolinkurl{meta-llama/Llama-3.2-3B-Instruct} \\
\texttt{deepseek\_v3\_1\_base} & \nolinkurl{deepseek-ai/DeepSeek-V3.1-Base} & \texttt{llama3\_3\_70B} & \nolinkurl{meta-llama/Llama-3.3-70B-Instruct} \\
\texttt{deepseek\_v3\_1} & \nolinkurl{deepseek-ai/DeepSeek-V3.1} & \texttt{ministral\_8B} & \nolinkurl{mistralai/Ministral-8B-Instruct-2410} \\
\texttt{gemma\_1B} & \nolinkurl{google/gemma-3-1b-it} & \texttt{mistral\_large} & \nolinkurl{mistralai/Mistral-Large-Instruct-2411} \\
\texttt{gemma\_4B} & \nolinkurl{google/gemma-3-4b-it} & \texttt{mistral\_small} & \nolinkurl{mistralai/Mistral-Small-Instruct-2409} \\
\texttt{gemma\_12B} & \nolinkurl{google/gemma-3-12b-it} & \texttt{olmo\_1B} & \nolinkurl{allenai/OLMo-2-0425-1B-Instruct} \\
\texttt{gemma\_27B} & \nolinkurl{google/gemma-3-27b-it} & \texttt{olmo\_13B} & \nolinkurl{allenai/OLMo-2-1124-13B-Instruct} \\
\texttt{gpt\_oss\_20B} & \nolinkurl{openai/gpt-oss-20b} & \texttt{olmo\_32B} & \nolinkurl{allenai/OLMo-2-0325-32B-Instruct} \\
\texttt{gpt\_oss\_120B} & \nolinkurl{openai/gpt-oss-120b} & \texttt{qwen3\_4B} & \nolinkurl{Qwen/Qwen3-4B-Instruct-2507} \\
\texttt{llama3\_1\_8B} & \nolinkurl{meta-llama/Meta-Llama-3.1-8B-Instruct} & \texttt{qwen3\_30B} & \nolinkurl{Qwen/Qwen3-30B-A3B-Instruct-2507} \\
\texttt{llama3\_1\_70B} & \nolinkurl{meta-llama/Meta-Llama-3.1-70B-Instruct} & \texttt{vicuna\_7B} & \nolinkurl{lmsys/vicuna-7b-v1.5} \\
\texttt{llama3\_2\_1B} & \nolinkurl{meta-llama/Llama-3.2-1B-Instruct} & \texttt{vicuna\_13B} & \nolinkurl{lmsys/vicuna-13b-v1.5} \\
\bottomrule
\end{tabularx}
\end{table*}

\section{Practical Recommendations}
\begin{itemize}\setlength{\itemsep}{2pt}
  \item \textbf{Schema-first authoring.} Keep IS/MSC as stable contracts; evolve content within those bounds to retain comparability across releases.
  \item \textbf{Trait leakage checks.} When expanding prompts, preserve the ``no explicit trait names/scores'' rule; regressions can inflate apparent performance under automatic readouts.
  \item \textbf{Culture portability.} For cross-cultural settings, duplicate MSC arenas with culture-specific norms and evaluate transfer before merging.
  \item \textbf{Adapter hygiene.} Store SFT and DPO adapters separately, with checksums and inference configs, to enable reproducibility and ablation rollbacks.
\end{itemize}

\FloatBarrier
\balance
\section{Licenses and Terms of Use}
\paragraph{External models.}
We used third-party base models (e.g., Llama, Vicuna, Qwen, Gemma, Mistral, DBRX, OLMo) strictly under their original licenses and model cards as cited in our references.
We do not redistribute these models; users must obtain them from their original providers and comply with the corresponding licenses.

\paragraph{External datasets/benchmarks.}
All external datasets/benchmarks cited in this work are used under their original licenses/terms as specified by their maintainers.
We do not redistribute any third-party data; users should access them from the official sources.

\paragraph{Our code.}
Project-original PsyAgent code is released under the Apache License 2.0 at \url{https://github.com/HKUST-FintechLab/PsyAgent}.
The repository at \url{https://github.com/Mengzibin/PsyAgent} is a read-only personal mirror.

\paragraph{Our artifacts.}
The synthetic IS$\times$MSC dataset, project-original documentation, and framework image are released under CC BY 4.0.
Third-party materials retain their original terms, and benchmark payloads whose redistribution rights were unclear are excluded.
Adapter weights are not included in the current public release; any future release will state the applicable terms and base-model restrictions.

\end{document}